\documentclass[11pt]{article}

\usepackage[T1]{fontenc}
\usepackage[utf8]{inputenc}
\usepackage[scaled=0.92]{helvet}

\usepackage{microtype}

\usepackage[
  top=2.2cm, bottom=2.6cm,
  left=2.0cm, right=2.0cm,
  headheight=1.4cm, footskip=1.2cm
]{geometry}

\usepackage{xcolor}
\definecolor{dfnavy}{RGB}{14,32,66}        
\definecolor{dfblue}{RGB}{0,62,126}        
\definecolor{dfteal}{RGB}{0,168,168}       
\definecolor{dftealdk}{RGB}{0,128,138}     
\definecolor{dfwarm}{RGB}{245,166,35}      
\definecolor{dfgray}{RGB}{86,99,118}       
\definecolor{dfbg}{RGB}{245,247,250}       
\definecolor{dfboxbg}{RGB}{235,243,247}    

\usepackage{multicol}
\usepackage{tcolorbox}
\tcbuselibrary{breakable, skins, raster}
\usepackage{titlesec}
\usepackage{enumitem}
\setlist[itemize]{leftmargin=1.1em, topsep=2pt, itemsep=2pt}
\setlist[enumerate]{leftmargin=1.4em, topsep=2pt, itemsep=2pt}
\usepackage{booktabs}
\usepackage{tabularx}
\usepackage{array}
\usepackage{multirow}
\usepackage{ragged2e}
\usepackage{amsmath, amssymb}
\usepackage{graphicx}
\usepackage{fancyhdr}
\usepackage{hyperref}
\hypersetup{
  colorlinks=true, urlcolor=dfteal, linkcolor=dfblue,
  pdftitle={Wellness Robots and the Path to Full Autonomy},
  pdfauthor={Mahoor, Hessler, Chileshe, Alsaadi, Emamian, Haring, Voyles}
}

\titleformat{\section}
  {\sffamily\bfseries\large\color{dfnavy}}
  {\thesection.}{0.6em}{\MakeUppercase}
  [\vspace{-2pt}{\color{dfteal}\rule{\linewidth}{1.2pt}}\vspace{-6pt}]

\titleformat{\subsection}
  {\color{dfblue}\sffamily\bfseries\normalsize}
  {\thesubsection}{0.6em}{}

\titleformat{\subsubsection}
  {\color{dftealdk}\sffamily\bfseries\small}
  {}{0em}{}

\titlespacing*{\section}{0pt}{14pt}{6pt}
\titlespacing*{\subsection}{0pt}{10pt}{3pt}
\titlespacing*{\subsubsection}{0pt}{8pt}{2pt}

\fancypagestyle{coverstyle}{
  \fancyhf{}
  
}

\newtcolorbox{keycallout}[1][]{
  enhanced, breakable,
  colback=dfboxbg, colframe=dfteal, boxrule=0pt,
  leftrule=3pt, leftrule=4pt,
  arc=0pt, outer arc=0pt,
  left=10pt, right=10pt, top=8pt, bottom=8pt,
  fontupper=\small,
  before skip=8pt, after skip=8pt, #1
}

\newtcolorbox{navycallout}[1][]{
  enhanced, breakable,
  colback=dfnavy, colframe=dfnavy, boxrule=0pt,
  arc=0pt, outer arc=0pt,
  left=12pt, right=12pt, top=10pt, bottom=10pt,
  coltext=white,
  fontupper=\small,
  before skip=10pt, after skip=10pt, #1
}

\newtcolorbox{warmcallout}[1][]{
  enhanced, breakable,
  colback=dfboxbg, colframe=dfwarm, boxrule=0pt,
  leftrule=4pt,
  arc=0pt, outer arc=0pt,
  left=10pt, right=10pt, top=8pt, bottom=8pt,
  fontupper=\small,
  before skip=8pt, after skip=8pt, #1
}

\newtcolorbox{figcol}[2]{%
  enhanced, breakable=false,
  colback=dfbg, colframe=dfteal, boxrule=0.5pt,
  arc=2pt, outer arc=2pt,
  left=10pt, right=10pt, top=10pt, bottom=10pt,
  before skip=8pt, after skip=8pt,
  fontupper=\small,
  title={\color{white}\sffamily\bfseries Figure #1},
  coltitle=white, colbacktitle=dfteal, fonttitle=\small\bfseries,
  attach boxed title to top left={xshift=8pt, yshift=-1pt},
  boxed title style={colback=dfteal, colframe=dfteal, sharp corners, boxrule=0pt, top=2pt, bottom=2pt, left=6pt, right=6pt},
  enlarge top initially by=4pt,
  height=5.5cm,
  #2
}

\newtcolorbox{figfull}[2]{%
  enhanced,
  colback=dfbg, colframe=dfteal, boxrule=0.5pt,
  arc=2pt, outer arc=2pt,
  left=12pt, right=12pt, top=12pt, bottom=12pt,
  before skip=10pt, after skip=10pt,
  fontupper=\small,
  title={\color{white}\sffamily\bfseries Figure #1},
  coltitle=white, colbacktitle=dfteal, fonttitle=\small\bfseries,
  attach boxed title to top left={xshift=10pt, yshift=-1pt},
  boxed title style={colback=dfteal, colframe=dfteal, sharp corners, boxrule=0pt, top=2pt, bottom=2pt, left=6pt, right=6pt},
  enlarge top initially by=5pt,
  height=6.5cm,
  #2
}

\newtcolorbox{figrealfull}[1]{%
  enhanced,
  colback=dfbg, colframe=dfteal, boxrule=0.5pt,
  arc=2pt, outer arc=2pt,
  left=12pt, right=12pt, top=12pt, bottom=12pt,
  before skip=10pt, after skip=10pt,
  fontupper=\small,
  title={\color{white}\sffamily\bfseries Figure #1},
  coltitle=white, colbacktitle=dfteal, fonttitle=\small\bfseries,
  attach boxed title to top left={xshift=10pt, yshift=-1pt},
  boxed title style={colback=dfteal, colframe=dfteal, sharp corners, boxrule=0pt, top=2pt, bottom=2pt, left=6pt, right=6pt},
  enlarge top initially by=5pt,
}

\newtcolorbox{figrealcol}[1]{%
  enhanced, breakable=false,
  colback=dfbg, colframe=dfteal, boxrule=0.5pt,
  arc=2pt, outer arc=2pt,
  left=10pt, right=10pt, top=10pt, bottom=10pt,
  before skip=8pt, after skip=8pt,
  fontupper=\small,
  title={\color{white}\sffamily\bfseries Figure #1},
  coltitle=white, colbacktitle=dfteal, fonttitle=\small\bfseries,
  attach boxed title to top left={xshift=8pt, yshift=-1pt},
  boxed title style={colback=dfteal, colframe=dfteal, sharp corners, boxrule=0pt, top=2pt, bottom=2pt, left=6pt, right=6pt},
  enlarge top initially by=4pt,
}

\newcommand{\figcaption}[1]{%
  \par\vspace{4pt}{\color{dfgray}\sffamily\small #1}\par
}

\newtcolorbox{dimensionbox}[1][]{
  enhanced, breakable,
  colback=white, colframe=dfteal, boxrule=0.6pt,
  arc=2pt, outer arc=2pt,
  left=10pt, right=10pt, top=8pt, bottom=8pt,
  fontupper=\small,
  before skip=8pt, after skip=8pt, #1
}

\renewcommand{\arraystretch}{1.25}

\newcommand{\dimensionheader}[2]{%
  \par\vspace{4pt}\noindent
  {\color{dfteal}\sffamily\bfseries#1}\hspace{0.5em}%
  {\color{dfgray}\sffamily\small\itshape#2}\par\vspace{2pt}%
}

\newcommand{\sensedecideact}[3]{%
  \begin{description}[leftmargin=!, labelwidth=4.2em, style=nextline, font=\bfseries\color{dfblue}\small, itemsep=1pt]
    \item[Sense] \small #1
    \item[Decide] \small #2
    \item[Act] \small #3
  \end{description}%
}

\begin{document}

\thispagestyle{empty}

\begin{flushleft}
{\color{dfgray}\sffamily\bfseries\small WHITE PAPER \;\textbullet\; PHYSICAL AI \& SOCIALLY ASSISTIVE ROBOTICS}
\end{flushleft}
\vspace{2pt}
{\color{dfteal}\rule{\linewidth}{0.8pt}}
\vspace{0.7cm}

\begin{center}
{\color{dfnavy}\sffamily\bfseries\fontsize{26}{32}\selectfont Wellness Robots and the\\Path to Full Autonomy\par}
\vspace{0.35cm}
{\color{dfgray}\sffamily\fontsize{14}{18}\selectfont A New Paradigm in AI-Powered Senior Care\par}
\vspace{0.25cm}
{\color{dfteal}\rule{3.2cm}{0.8pt}}\par
\vspace{0.25cm}
{\color{dfgray}\sffamily\itshape\fontsize{10.5}{14}\selectfont Defining the Autonomous Wellness Robot:\\
Wellness Dimensions, the CRAS Framework, and the Roadmap\\
from Current Practice to Full Autonomy\par}
\end{center}

\vspace{0.55cm}

\begin{center}
{\sffamily\normalsize
Mohammad H.\ Mahoor, Ph.D.$^{1,*}$ \quad
Matthew Hessler$^{1,2}$ \quad
Lombe Chileshe$^{1,2}$ \quad
Ola Alsaadi$^{2}$\\[3pt]
Eshrat Emamian, Ph.D.$^{2,\dagger}$ \quad
Kerstin S.\ Haring, Ph.D.$^{1}$ \quad
Richard M.\ Voyles, Ph.D.$^{3}$\par}
\end{center}
\vspace{0.2cm}

\begin{center}
\small\sffamily
\begin{tabular}{rl}
$^{1}$ & Ritchie School of Engineering and Computer Science, University of Denver, Denver, Colorado, USA\\
$^{2}$ & DreamFace Technologies LLC, Denver, Colorado, USA\\
$^{3}$ & Department of Engineering, University of Texas at Arlington, Arlington, Texas, USA\\
\end{tabular}\\[3pt]
{\color{dfgray}\itshape $^{*}$Founder, DreamFace Technologies LLC. \quad $^{\dagger}$Co-founder, DreamFace Technologies LLC.}
\end{center}

\vspace{0.5cm}

\noindent{\color{dfnavy}\sffamily\bfseries\normalsize ABSTRACT}\hspace{8pt}{\color{dfteal}\rule[2pt]{\dimexpr\linewidth-3cm\relax}{0.6pt}}\par
\vspace{3pt}
{\small This paper introduces the \textbf{autonomous wellness robot} as a distinct category of socially assistive system, defined by its capacity to continuously assess and respond across the seven ICAA dimensions of wellness (see Figure~\ref{fig:2}). We propose the \textbf{Care Robot Autonomy Scale (CRAS)}: a six-level taxonomy (L0--L5) modeled on the automotive automation standard SAE~J3016 and evaluated across four wellness-aligned dimensions. We present \textbf{Ryan} — developed at the University of Denver and commercialized by DreamFace Technologies — as a case study. NIH/NIA-funded longitudinal evidence ($N{=}18$; 887~hours) shows statistically significant cognitive improvement (SLUMS $+2.61$, $p{=}0.015$) and a clinically meaningful directional reduction in depressive symptoms (PHQ-9 $-2.50$, $p{=}0.073$), with no novelty effect. We position Ryan\textsuperscript{\textregistered} today at CRAS Level~1--2, with emerging Level~3 capabilities in some dimensions, and outline a three-phase roadmap to L5 (full autonomy) by the early 2030s.\par}

\vspace{6pt}
\noindent{\color{dfnavy}\sffamily\bfseries\normalsize INDEX TERMS}\hspace{8pt}{\color{dfteal}\rule[2pt]{\dimexpr\linewidth-3.2cm\relax}{0.6pt}}\par
\vspace{3pt}
{\small\sffamily Socially assistive robotics, wellness robots, autonomous systems, senior care, dementia, Physical AI, vision--language--action models, CRAS, SAE J3016, NIH/NIA, FHIR.}

\vfill
{\color{dfteal}\rule{\linewidth}{0.6pt}}\par
\vspace{3pt}
\noindent{\color{dfgray}\sffamily\small\textit{May 2026}}\hfill{\color{dfteal}\sffamily\bfseries\small dreamfacetech.com}

\newpage

\refstepcounter{figure}\begin{figrealfull}{1. The Ryan\textsuperscript{\textregistered} Wellness Robot}
\label{fig:1}
\begin{center}
\includegraphics[height=11cm,keepaspectratio]{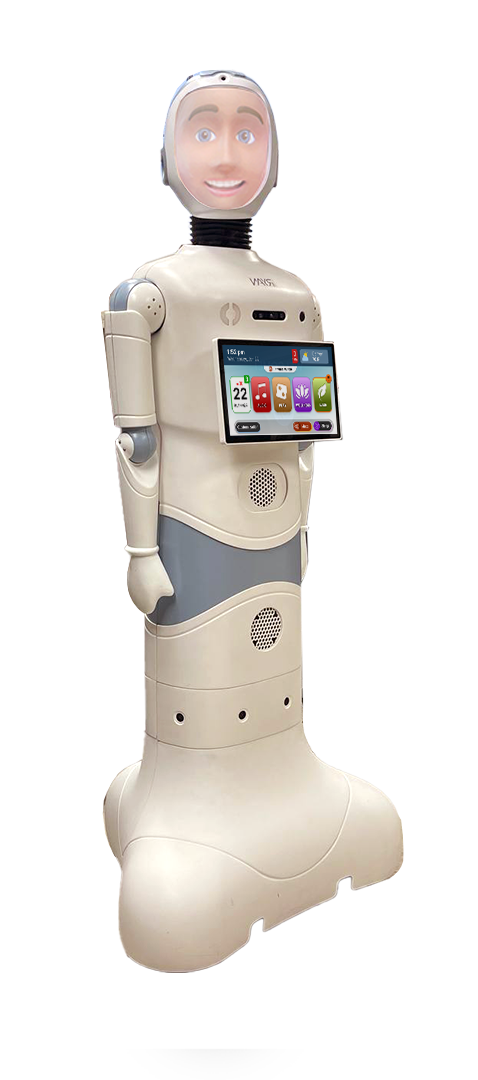}
\end{center}
\figcaption{Ryan, the AI-powered socially assistive wellness robot developed at the University of Denver and commercialized by DreamFace Technologies. Ryan combines an animated face with multimodal perception, conversational AI, and a touchscreen interface to deliver engagement across the seven dimensions of wellness in senior care settings.}
\end{figrealfull}

{\color{dfnavy}\sffamily\bfseries\Large EXECUTIVE SUMMARY\par}
\vspace{2pt}
{\color{dfteal}\rule{\linewidth}{1.2pt}}\par
\vspace{4pt}

\begin{multicols}{2}

The senior care system is in crisis. By 2050, the United States will face 16~million people living with Alzheimer's disease and related dementias (AD/ADRD). 60\% of senior care facilities fear closure due to staffing shortages. The average assisted-living resident receives fewer than 18~minutes of structured wellness programming per day. Loneliness affects 56\% of residents broadly and rises to 72\% among those with cognitive impairment. Traditional care models cannot scale to meet this demand.

This paper proposes a different paradigm: \textbf{wellness} — the holistic, multi-dimensional cultivation of a purposeful life — as the organizing principle for a new category of socially assistive robot. The International Council on Active Aging defines wellness across seven dimensions: \textbf{Physical, Emotional, Social, Intellectual, Spiritual, Environmental, and Occupational}. An autonomous \textit{wellness robot} is not a generic care robot, a companion device, or a medical device — it is a system whose behavior is defined by its capacity to continuously assess and respond across all seven dimensions, with longitudinal memory, dignity-preserving interaction, and care-team escalation built in by design.

\columnbreak

To make this category measurable, we introduce the \textbf{Care Robot Autonomy Scale (CRAS)} — a six-level taxonomy (L0–L5) modeled on the automotive automation standard SAE J3016 and evaluated across four wellness-aligned dimensions: assessment, intervention, social intelligence, and care coordination. We then enumerate the seven technical capabilities required to climb this scale: multimodal sensing, adaptive planning, personalization with longitudinal memory, dignity-preserving safety, privacy and data sovereignty, care-team integration, and explainable decisions.

We present \textbf{Ryan} (see Figure~\ref{fig:1}), developed and commercialized by DreamFace Technologies, as a case study showing that the autonomous wellness robot category is already emerging. NIH/NIA-funded longitudinal evidence (R44AG059483; 18~participants; 887~hours) shows a statistically significant improvement in cognitive function (SLUMS mean change $+2.61$, $p=0.015$, $d_z=0.64$) and a clinically meaningful directional improvement in depression (PHQ-9 mean change $-2.50$, $p=0.073$, $d_z=0.45$), with no novelty effect. A robot-led yoga study \cite{mapoles2025yoga} reports significant balance ($p=0.046$, $d=-0.853$) and cognition ($p=0.032$) improvement.

We position Ryan today at \textbf{CRAS Level~1--2} (see Figure~\ref{fig:3}) with emerging Level~3 capabilities in some dimensions, and describe the roadmap toward L3, L4, and L5. The paper closes with implications for care operators, researchers, regulators, and robot platform developers.

\end{multicols}

\vspace{4pt}
\begin{tcolorbox}[enhanced, colback=dfnavy, colframe=dfnavy, boxrule=0pt, arc=0pt,
  left=14pt, right=14pt, top=10pt, bottom=10pt, coltext=white]
\sffamily\small
\begin{tabularx}{\textwidth}{@{}>{\color{dfteal}\bfseries}l X@{}}
\textsc{The crisis} & 60\% of facilities fear closure; 16M~Americans projected with AD/ADRD by 2050; \$1T+ annual costs.\\
\textsc{The gap} & $<$18~minutes of structured wellness per resident per day; 56--72\% report loneliness.\\
\textsc{The category} & Autonomous wellness robot $\neq$ companion device, medical device, or general-purpose care robot.\\
\textsc{The scale} & CRAS L0--L5: a six-level autonomy framework modeled on SAE~J3016, evaluated across 4 dimensions.\\
\textsc{The evidence} & SLUMS +2.61 ($p{=}0.015$), PHQ-9 $-2.50$ ($p{=}0.073$); no novelty effect over 887 hours of in-home use (NIH/NIA R44AG059483).\\
\textsc{The roadmap} & Phased technical milestones from current L1--2 toward L3, L4 (agentic), and L5 (full autonomy).\\
\end{tabularx}
\end{tcolorbox}

\begin{multicols}{2}

\section{The Senior Wellness Crisis: Why Traditional Care Models Cannot Scale}

The convergence of demographic, workforce, and economic pressures has produced a senior care system that is unable to meet the needs of the people it serves. Understanding why traditional models cannot scale is the prerequisite for understanding why a new paradigm — and a new category of robot — is needed.

\subsection{The Demographic Reality}

The global population aged~65 and over is projected to more than double by 2050, reaching approximately 1.6~billion. In the United States alone, the Alzheimer's Association estimates that the population living with Alzheimer's disease and AD-related dementias (AD/ADRD) could reach 16~million by mid-century \cite{alzAssoc2024}, with annual direct care costs exceeding \$1~trillion. AD/ADRD is not the only driver: depression affects roughly 30--40\% of older adults in long-term care, while loneliness — independently associated with accelerated cognitive decline, depression, and mortality — has reached pandemic proportions. The COVID-19 era exposed how thin the margins are: facility-wide social distancing, already against a backdrop of severe staffing shortages, transformed loneliness from a quality-of-life concern into a measurable mortality risk factor.

\subsection{The Workforce Crisis}

The supply side of senior care has been deteriorating for over a decade and is now in critical condition. PHI National projects a shortfall of roughly 1.2~million direct care workers in the United States by 2030 \cite{phi2024}. A 2024 LeadingAge survey of 759 facility operators found that \textbf{60\% feared closure} \cite{leadingAge2024} due to staffing pressures. Median direct-care wages remain near minimum wage in many states, while turnover rates of 60--100\% per year compound the problem: each new caregiver requires onboarding, builds resident relationships from scratch, and is statistically likely to be gone within a year. The result is a workforce structurally unable to deliver the relational, longitudinal care that older adults — and especially those with cognitive impairment — require.

\subsection{The Engagement Gap}

This workforce reality produces a measurable engagement gap inside the facility itself. The average assisted-living resident receives \textbf{fewer than 18~minutes} of structured wellness programming per day. The typical activities staff-to-resident ratio is approximately \textbf{1:47}, which makes systematic individual programming structurally impossible. Group activity formats — bingo, sing-alongs, exercise classes — are inherently constrained: they happen at fixed times, address two or three wellness dimensions at most, and are not personalized. Residents who decline to attend, sleep through morning sessions, or simply prefer one-on-one interaction are largely unserved. The 16+~hours each day outside of programming time — evenings, weekends, holidays — are essentially uncovered. For residents with cognitive impairment, this lack of structure is itself a clinical risk factor.

\subsection{Why Healthcare Automation Alone Is Not the Answer}

Several technology categories have been positioned, by various vendors, as solutions to this crisis: medication reminder apps, telehealth platforms, fall-detection wearables, voice assistants, tablet-based engagement programs, and emotional companion robots. Each addresses some piece of the problem. None, individually or collectively, addresses the central issue. The reason is that the senior care crisis is not primarily a \textit{healthcare automation} problem. Reminding someone to take a pill, escalating a fall, or asking ``how are you feeling today?'' on a tablet are point interventions; they do not in themselves constitute care. \textit{Healthcare automation is not wellness autonomy.} A reminder is not a relationship; a detection is not an intervention; a generic conversation prompt is not adaptive engagement. A robot that delivers any one of these in isolation is a useful appliance, not an autonomous wellness robot.

What is required is something different in kind: a system that \textit{continuously} assesses an individual's state across multiple dimensions of wellbeing, \textit{adaptively} plans interventions, \textit{remembers} who the person is over weeks and months, \textit{escalates} appropriately when concerns are detected, and does so while \textit{respecting} the resident's autonomy and dignity. That is not automation. That is a new category of system, requiring its own paradigm, its own standards, and its own measurable autonomy framework. Establishing those is the project of this paper.

\end{multicols}

\begin{multicols}{2}

\section{Wellness as the Central Paradigm: The Seven Dimensions}

The conceptual foundation for the autonomous wellness robot is not robotics. It is wellness — specifically, the multi-dimensional model articulated by the International Council on Active Aging (ICAA) \cite{icaa7dim} and adopted broadly across the senior living industry. The ICAA model defines wellness as \textit{the capacity to lead a purposeful and engaged life}, supported across seven interdependent dimensions: \textbf{Physical, Emotional, Social, Intellectual, Spiritual, Environmental,} and \textbf{Occupational}. We argue that this seven-dimension framework is not merely a useful taxonomy — it is the criterion set against which any candidate wellness robot should be evaluated.

In what follows, we describe each dimension along four lines: \textit{what it means} in senior care, \textit{why it matters} clinically, \textit{the gap} in current human-delivered care, and \textit{what an autonomous robot would need to sense, decide, and do} to support it. These four-part dimension descriptions become, in Section~5, the source from which we derive the technical capabilities required for autonomy, and in Section~4, the basis for the CRAS evaluation rubric.

\end{multicols}

\refstepcounter{figure}\begin{figrealfull}{2. The Seven Dimensions of Wellness (ICAA)}
\begin{center}
\label{fig:2}
\includegraphics[width=0.78\linewidth,keepaspectratio]{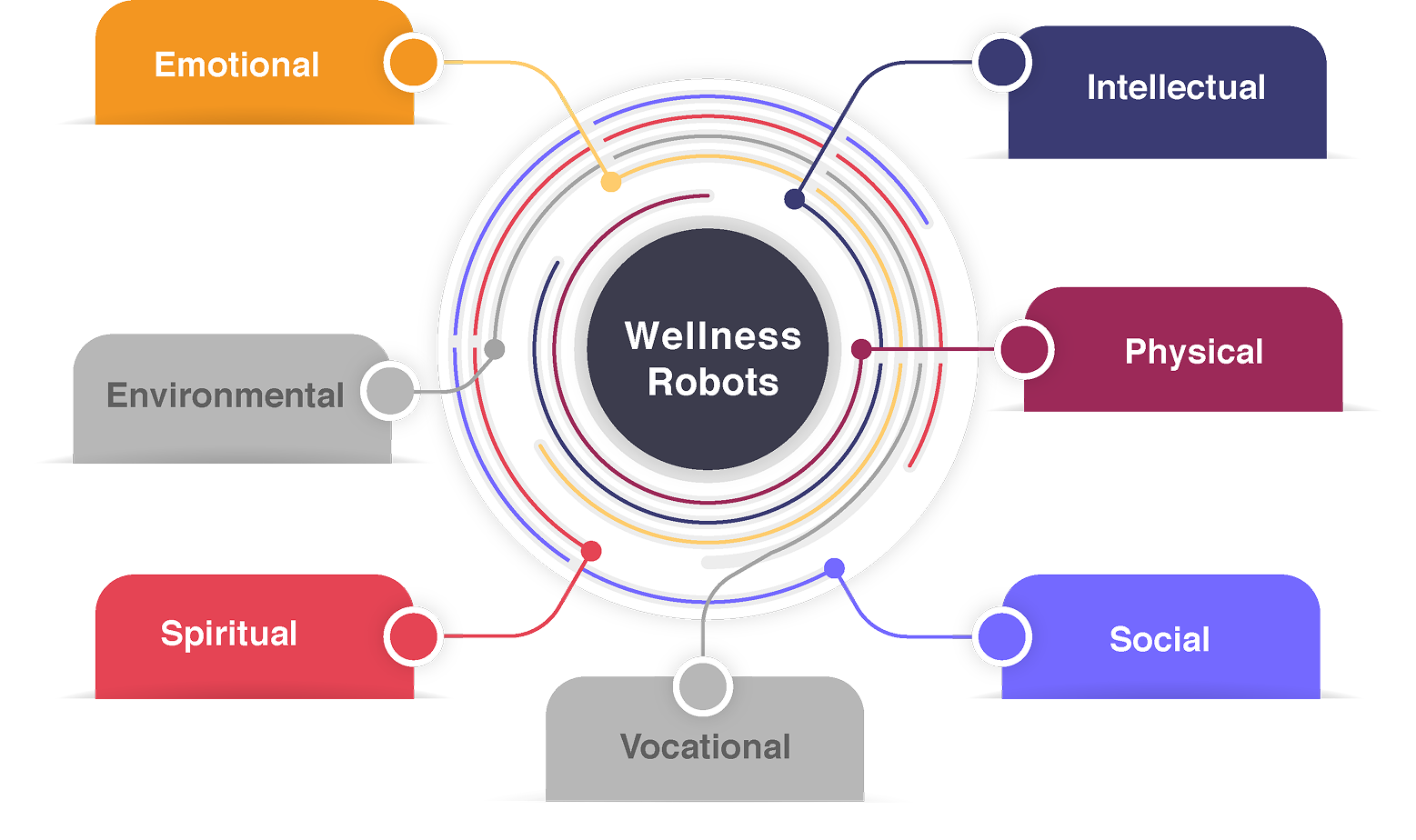}
\end{center}
\figcaption{The seven interconnected dimensions of wellness — Physical, Emotional, Social, Intellectual, Spiritual, Environmental, and Occupational — form the conceptual foundation for the autonomous wellness robot category. Source: International Council on Active Aging; image courtesy of DreamFace Technologies.}
\end{figrealfull}

\begin{dimensionbox}
\dimensionheader{1. Physical Wellness}{The body in motion: strength, balance, mobility, and the activities of daily living.}

\textbf{In senior care.}\; Physical wellness has two faces. The \textit{engagement} side covers regular exercise, mobility, nutrition, sleep, and sensory health — even modest activity (chair exercise, gentle yoga, balance training) reduces fall risk and slows physical decline. The \textit{assistance} side covers the activities of daily living (ADLs) when independent performance becomes hard: transfers, ambulation, fall response, dressing, eating. The wellness paradigm holds both: a flourishing later life requires being supported in motion, not just instructed in motion. \\[2pt]
\textbf{Why it matters clinically.}\; Falls are the leading cause of injury-related death in adults over 65 — the CDC estimates one in four older adults falls each year \cite{cdcFalls2024}. Sedentary behavior accelerates cognitive decline, depression, and mortality. ADL dependence is the strongest single predictor of nursing-home placement, and unmet ADL need correlates directly with depression and hospitalization risk. \\[2pt]
\textbf{The gap.}\; Group exercise classes occur 3--5 times per week; physical and occupational therapy visits are weekly at most; evenings, weekends, and holidays are largely uncovered. On the assistance side, residents who need help to stand, ambulate, or recover from a near-fall typically wait for a staff member who is responding to multiple call lights. Both gaps are structural to the staffing crisis and cannot be closed by adding more group programming. \\[2pt]
\sensedecideact
{Vision-based skeletal tracking; pose, gait, balance kinematics; sleep/activity rhythms; observed safety signals (e.g., walker use, near-falls, prolonged sedentary time, posture transitions).}
{Adapt exercise difficulty and modality to today's energy and joint range; identify decreasing balance scores trending toward fall risk; decide when an exercise should be halted; decide whether an observed event requires staff alert vs.\ on-robot physical response.}
{Lead chair exercise, yoga, and passive-motion sessions; provide form feedback in real time; log objective range-of-motion and balance metrics; escalate concerning trends to PT/OT; at higher CRAS levels, provide direct physical assistance — sit-to-stand support, ambulation guidance, fall response — within force and dignity envelopes (see Section~5.5).}
\end{dimensionbox}

\begin{dimensionbox}
\dimensionheader{2. Emotional Wellness}{Recognizing, expressing, and regulating emotions.}

\textbf{In senior care.}\; Emotional wellness covers mood, self-awareness, the ability to express feelings, and access to empathic support. It is fundamentally relational — most adults regulate emotion through conversation with someone who listens. \\[2pt]
\textbf{Why it matters clinically.}\; Depression in older adults is underdiagnosed and undertreated; up to 30--40\% of long-term care residents meet criteria. Untreated depression accelerates AD/ADRD progression, worsens chronic disease outcomes, and is strongly associated with suicide risk in adults over 75. \\[2pt]
\textbf{The gap.}\; Most residents have access to a clinical psychologist for less than one hour per month. Day-to-day emotional support depends on the time and skill of caregivers, who in understaffed environments rarely have minutes to spare for sustained empathic conversation. \\[2pt]
\sensedecideact
{Multimodal affect recognition (face, voice prosody, language sentiment); discrepancies between verbal and nonverbal channels; longitudinal mood trajectories.}
{When to listen vs.\ redirect; when to lighten with humor; when an observed pattern crosses a clinical threshold for escalation; when to administer a structured screen (PHQ-9, GAD-7).}
{Empathic conversation in real time; facial-expression mirroring; mood-congruent music and topics; structured PHQ-9 screening; flag clinically meaningful change to nursing.}
\end{dimensionbox}

\begin{multicols}{2}

\subsection*{The interdependence of dimensions}

\noindent A central insight of the ICAA framework \cite{icaa7dim} is that the seven dimensions are not parallel silos. Sleep (Physical) shapes mood (Emotional). Loneliness (Social) accelerates cognitive decline (Intellectual). Loss of purpose (Occupational) often presents clinically as depression (Emotional). A care system that addresses any one dimension in isolation — as nearly all current technologies do — will systematically miss the cross-dimensional dynamics that make up the lived experience of a resident.

This interdependence is the single most important reason why an autonomous wellness robot cannot be assembled by stitching together a fall detector, a medication reminder, a conversational chatbot, and a video-call interface. Each component would optimize one dimension while ignoring the others. The architecture of the wellness robot must, from its first design decision, treat the dimensions as a coupled system.

\end{multicols}

\begin{dimensionbox}
\dimensionheader{3. Social Wellness}{Meaningful connection and a sense of belonging.}

\textbf{In senior care.}\; Social wellness is built from regular conversation, friendship, family contact, shared activity, and a sense of being known by other people. Loss of social ties is often the silent cost of moving into care. \\[2pt]
\textbf{Why it matters clinically.}\; Social isolation in older adults carries mortality risk comparable to smoking 15~cigarettes per day \cite{holtLunstad2010}. It accelerates cognitive decline at rates similar to a clinical APOE risk allele. The U.S.\ Surgeon General has formally declared loneliness a public health epidemic \cite{surgeonGeneral2023}. \\[2pt]
\textbf{The gap.}\; Residents typically have one-on-one engagement with staff for only 10--20~minutes per day. Family contact may be weekly or less. Even residents living among 50~other people can spend most of their day alone in their apartment. \\[2pt]
\sensedecideact
{Conversation frequency, depth, and reciprocity; whether the resident initiates exchanges; references to family and past relationships; signs of withdrawal.}
{When to engage in companionable conversation, when to facilitate communication within the resident's social network, when isolation has crossed a clinical threshold, when to recommend a community activity.}
{Conduct natural, multi-turn, personalized conversation; remember details across sessions; enable connections across the resident's social network — family, friends, faith community, and peers — through voice, video, and in-person bridging.}
\end{dimensionbox}

\begin{dimensionbox}
\dimensionheader{4. Intellectual Wellness}{Curiosity, learning, and cognitive engagement.}

\textbf{In senior care.}\; Intellectual wellness is sustained through novel ideas, problem-solving, reading, games, and challenging conversation. The brain follows the same ``use it or lose it'' principle that applies to muscles. \\[2pt]
\textbf{Why it matters clinically.}\; Cognitive Stimulation Therapy (CST) — group sessions that exercise memory, attention, and reasoning — produces measurable cognitive improvement in adults with mild-to-moderate dementia \cite{woods2012cst} ($g=0.41$). Sustained intellectual engagement is among the strongest modifiable factors in dementia trajectory. \\[2pt]
\textbf{The gap.}\; CST requires a trained facilitator and a group setting; many facilities cannot offer it. Even where available, it reaches only the small subset of residents who attend group programs. Homebound and severely impaired residents are systematically excluded. \\[2pt]
\sensedecideact
{Performance on adaptive cognitive games; conversation complexity over time; longitudinal SLUMS and BIMS scores; topic-specific engagement signals.}
{Adapt game difficulty in real time; choose between memory, executive function, and language tasks based on recent performance; identify performance decline trends.}
{Deliver one-on-one CST-equivalent activities on demand; administer SLUMS and BIMS in conversational form; report longitudinal trends to clinical team.}
\end{dimensionbox}

\begin{dimensionbox}
\dimensionheader{5. Spiritual Wellness}{Meaning, values, and the inner life.}

\textbf{In senior care.}\; Spiritual wellness encompasses religious practice, meditation, life review, gratitude practice, the sense of being part of something larger than oneself, and existential questions of meaning that often arise toward end of life. \\[2pt]
\textbf{Why it matters clinically.}\; A robust evidence base associates regular spiritual practice with reduced depression, improved coping with chronic illness, and better adjustment to end-of-life care. For residents whose lifelong faith communities are no longer accessible due to mobility limitations, the loss can be profound. \\[2pt]
\textbf{The gap.}\; Many facilities offer non-denominational chapel service once per week. Residents from minority faiths, those who prefer private practice, and those with attendance barriers are largely underserved. \\[2pt]
\sensedecideact
{Stated faith preferences; conversational references to meaning, family legacy, and end-of-life concerns; observed patterns suggesting spiritual distress vs.\ engagement.}
{Respect autonomy: never proselytize; honor stated tradition; recognize when a topic warrants pastoral referral; identify when reminiscence is therapeutic vs.\ avoidant.}
{Lead Bible study, the Rosary, guided meditation, or gratitude practice on resident's terms; play sacred music; facilitate life review; refer to chaplaincy when appropriate.}
\end{dimensionbox}

\begin{dimensionbox}
\dimensionheader{6. Environmental Wellness}{Safety, comfort, and the surrounding physical world.}

\textbf{In senior care.}\; Environmental wellness covers the physical environment a person lives in: lighting, temperature, clutter, accessibility, fall hazards, sensory inputs (noise, scent), and the broader question of feeling at home in one's space. \\[2pt]
\textbf{Why it matters clinically.}\; Environmental hazards are the precipitating cause of most falls. Poor lighting accelerates sundowning in dementia. Cluttered or unfamiliar spaces drive disorientation and agitation. Conversely, a familiar, well-arranged room is itself therapeutic. \\[2pt]
\textbf{The gap.}\; Environmental assessment is typically a once-per-admission OT activity. Day-to-day monitoring of changing conditions — a misplaced rug, a dimmed bulb, a thermostat off-target — depends on caregivers noticing during brief check-ins. \\[2pt]
\sensedecideact
{Ambient lighting, temperature, sound; obstacle detection during navigation; longitudinal mapping of clutter and rearrangement; cross-references to recent fall or near-miss events.}
{Differentiate ``new arrangement'' from ``hazard''; respect resident autonomy over their personal space; decide what is reportable to facilities team vs.\ resident.}
{Notify resident or staff of detected hazards; adjust ambient music, video, or lighting (where integrated) to support sundowning protocols; log environmental context with each adverse event.}
\end{dimensionbox}

\begin{dimensionbox}
\dimensionheader{7. Occupational Wellness}{Purpose, contribution, and the use of one's skills.}

\textbf{In senior care.}\; Occupational wellness in older adults encompasses the use of lifelong skills, continued contribution to family and community, hobbies, mentorship, volunteer work, and the basic human need to feel useful. ``Retired'' is not the same as ``without purpose.'' \\[2pt]
\textbf{Why it matters clinically.}\; Loss of purpose is among the strongest predictors of depression in older adults. Erikson's developmental framework \cite{erikson1950} names ``generativity vs.\ stagnation'' as the central late-life psychological task. Residents who feel useful demonstrate measurably better mood, cognition, and physical activity. \\[2pt]
\textbf{The gap.}\; Few facilities have systematic programs for engaging residents' lifelong skills. The retired engineer, teacher, surgeon, or musician is generally treated as a recipient of activities rather than a contributor. \\[2pt]
\sensedecideact
{Conversational references to past work, hobbies, family roles; expressed wishes to contribute; observed engagement vs.\ disengagement during purpose-related activities.}
{Identify opportunities to reflect, teach, or mentor; calibrate cognitive load to current capacity; recognize when reminiscence is generative vs.\ depressive rumination.}
{Support hobbies (gardening, music, crafts) and skill-based projects; facilitate intergenerational mentorship sessions via video; conduct structured life-review and legacy-recording activities.}
\end{dimensionbox}

\begin{multicols}{2}

\subsection*{From dimensions to autonomy criteria}

The seven-dimension framework gives us, for the first time, a principled basis for evaluating what an autonomous wellness robot must do. A system that supports two or three dimensions in narrow ways is a useful product, but it is not in the same category as one that addresses all seven continuously. A system that performs scripted activities in each dimension is at a different autonomy level than one that adapts in real time and remembers across sessions. A system that operates on its own initiative across all seven dimensions, escalates appropriately, integrates with the clinical team, and maintains dignity-preserving behavior is what we will define in Section~4 as a CRAS Level~5 wellness robot.

We turn next to the categorical question that must be answered before autonomy can be measured: what makes a robot a wellness robot in the first place?

\end{multicols}

\begin{multicols}{2}

\section{What Makes a Robot a Wellness Robot}

The term ``wellness robot'' is, today, used loosely. Marketing materials apply it to companion plush toys, animated tablets, programmable humanoids, voice-only smart speakers, and exoskeletons. If the category is to mean anything — for clinicians choosing technologies, for regulators writing certification rules, for researchers comparing systems — it must be defined.

We define an \textbf{autonomous wellness robot} as a socially assistive robot \cite{feilSeifer2005sar} whose design, behavior, and evaluation are organized around continuous engagement across the seven dimensions of wellness, with longitudinal memory of the individuals it serves, dignity-preserving interaction, and built-in care-team escalation. This definition is meant to be discriminating: it places the wellness robot in its own category, distinct from neighboring categories that share some of its components.

\subsection{What a Wellness Robot Is Not}

\subsubsection{Not a companion device.}
Companion devices like PARO (the seal-shaped therapeutic robot) \cite{wada2003paro} provide tactile and emotional comfort through a single modality. The clinical evidence for PARO is real — meta-analytic effect sizes of $-0.40$ on depression and $-0.27$ on agitation in dementia \cite{rashid2024paro}. But PARO does not assess. It does not adapt across dimensions. It does not remember. It does not escalate. It is a beautifully designed therapeutic appliance whose value lies in what it elicits from the human, not in what it understands or decides. Companion devices are valuable. They are not wellness robots.

\subsubsection{Not a medical device.}
Medical devices — fall detectors, gait analyzers, automated medication dispensers, blood pressure monitors — are designed to perform a specific clinical measurement or task with regulated reliability. Their FDA pathway, design priorities, and operational scope are all narrow and clinical. A wellness robot may incorporate medical-grade sensing (and may itself become regulated as Software as a Medical Device for specific functions). But its primary purpose is not to measure a single clinical variable; it is to support human flourishing across multiple dimensions over time.

\subsubsection{Not a general-purpose care robot.}
Generalist humanoid platforms — Unitree, Tesla Optimus — are designed for broad physical manipulation and locomotion. They are formidable engineering achievements, but they are not wellness robots out of the box. Without an integrated wellness sensing stack, a multi-dimensional intervention library, longitudinal personalization, dignity-aware behavior, and clinical escalation pathways, a humanoid in a memory-care unit is a powerful general-purpose machine that happens to be in a senior care facility — not a wellness robot. The mistake to avoid is treating the platform as the product. The platform is necessary; the wellness software, data, and clinical integration are what define the category.

\subsubsection{Not a chatbot or voice assistant.}
Tablet-based engagement programs and voice assistants offer language interaction, scheduled reminders, and increasingly capable conversation. They lack embodiment, multimodal affect perception (face and prosody, not just text), physical activity guidance, and most critically, the relational grounding that comes from a present, embodied entity that recognizes you across days and weeks.

\subsection{Eight Defining Properties of a Wellness Robot}

We propose eight properties that, taken together, define the autonomous wellness robot. These properties also become the source for the technical capability requirements in Section~5 and the evaluation criteria for the CRAS framework in Section~4.

\textbf{(1) Multi-dimensional coverage.} The system supports interventions across all seven ICAA dimensions, not a subset. It is structurally designed for cross-dimensional reasoning — observing in one dimension and acting in another (e.g., detecting low mood and adjusting today's exercise, music, and conversation simultaneously).

\textbf{(2) Continuous, ambient assessment.} The system continuously and unobtrusively assesses wellness state during ordinary interaction. It does not require the resident to formally enter ``assessment mode.'' The assessment is multimodal: face, prosody, language, behavior, and physical movement, integrated over time.

\textbf{(3) Adaptive intervention planning.} Sessions are not scripted. The system selects and shapes interventions based on the resident's current state, longitudinal history, stated preferences, and clinical context. The same person on two different mornings receives two different sessions.

\textbf{(4) Longitudinal personalization.} The system maintains a per-resident profile that grows over time: family members named, preferred topics, medical context, faith tradition, daily rhythms, conversational quirks. This memory is what allows the relationship to deepen rather than reset each day.

\textbf{(5) Dignity-preserving interaction.} The system is designed around resident autonomy. It does not surveil; it engages. It does not infantilize. It respects refusal. It explains itself in plain language. It maintains confidence without the brittleness that turns interaction into power struggle.

\textbf{(6) Embodied social presence.} The system is embodied — physically present, with a face, voice, and body that residents can locate, recognize, and form a relationship with. Embodiment is not cosmetic; it is what enables attachment, trust, and the felt sense of being accompanied that the literature on socially assistive robotics consistently identifies as therapeutically distinct from screen-based interaction.

\textbf{(7) Care-team integration and escalation.} The system is part of the care team, not a parallel channel. It writes structured observations to the care record, raises clinical flags when appropriate, defers to humans for decisions outside its scope, and amplifies — rather than replaces — the work of caregivers, activities staff, therapists, and clinicians.

\textbf{(8) Privacy and data sovereignty.} The system is designed around the principle that resident data belongs to the resident. Assessment scores (PHQ-9, SLUMS, GAD-7) are the resident's clinical record; the operator does not collect them as a business asset. Conversational content is processed on-device or under strict access control. Federated and on-device learning architectures prevent unnecessary cloud egress.

\begin{navycallout}
\textbf{The defining test.} A system is an autonomous wellness robot to the extent that it satisfies all eight properties. A system that satisfies most of them is on the path. A system that satisfies only one or two — however excellent in that domain — belongs to a different category.
\end{navycallout}

\subsection{The Difference Between Healthcare Automation and Wellness Autonomy}

Healthcare automation reduces the labor cost of a defined task: dispense the pill, document the vital sign, schedule the appointment, transcribe the note. These are real and valuable contributions. But they share a structural feature: each is a discrete, well-specified function with a clear input and output.

Wellness autonomy is structurally different. The ``input'' is a person — multi-dimensional, longitudinal, contradictory. The ``output'' is a relationship and a set of small daily decisions whose aggregate effect, over months, is a flourishing or declining life. There is no single function to specify. The system must hold many goals at once, balance them in context, change its mind as the person changes, and report what it has learned to humans whose own judgment it complements rather than replaces.

This is why the wellness robot category cannot be reached by adding more automated tasks to a tablet, or by attaching a wellness app to a humanoid platform. It requires designing from the wellness paradigm down — and that is also why it requires its own autonomy framework, which we develop next.

\end{multicols}

\begin{multicols}{2}

\section{CRAS: Defining Autonomy in the Wellness Context}

The Society of Automotive Engineers (SAE) defined six levels of driving automation (J3016, L0--L5) \cite{saeJ3016} that have become the global lingua franca for autonomous vehicles. The framework's contribution was not the levels themselves; it was making automation \textit{discussable} — giving engineers, regulators, insurers, and the public a shared vocabulary for what a system can and cannot do, and what it requires from a human.

The senior care field needs an equivalent. We propose the \textbf{Care Robot Autonomy Scale (CRAS)} — a six-level framework (L0--L5) modeled on J3016 and adapted for the wellness paradigm.

\subsection{Prior Autonomy Frameworks and What They Miss}

CRAS does not appear in a vacuum; the field has produced several relevant autonomy frameworks. Understanding what they cover — and what they leave out — clarifies why a wellness-specific scale is needed.

\textbf{SAE J3016} \cite{saeJ3016} defines driving automation across six levels (L0--L5) along a single dimension (driving) and a clean human-vs-machine division of responsibility. It is the most successful autonomy framework yet published, but is structurally tied to a single, well-defined task with a clear failure mode (a crash).

\textbf{NIST ALFUS} (Autonomy Levels for Unmanned Systems) \cite{huangAlfus2008} extended the autonomy-levels idea to unmanned vehicles by combining three axes: mission complexity, environmental difficulty, and human-independence. ALFUS is closer to the wellness setting in acknowledging that autonomy is multi-dimensional, but it remains task-and-mission-oriented rather than relationship-and-care-oriented.

\textbf{Beer, Fisk, and Rogers' Levels of Robot Autonomy in HRI} \cite{beerFiskRogers2014} is the most directly relevant precursor: a ten-level scale spanning teleoperation through full autonomy, evaluated against the human-robot \textit{interaction} loop. It makes the critical move of placing the human in the loop of the autonomy definition. However, its scope is general HRI rather than care-specific, and it offers no rubric for evaluating clinical assessment, intervention delivery, or care-team integration as autonomy dimensions in their own right.

\textbf{Healthcare-AI maturity frameworks} \cite{ukNhsAiMaturity2024, fdaSamd2022} have begun to appear in regulatory contexts, but they treat AI as a software system rather than as an embodied, social, longitudinal care partner. They do not engage with the wellness dimensions, social intelligence, or the dignity-preservation requirements that distinguish wellness robots from medical software.

\textbf{The gap.} What is missing across these frameworks is a scale that (a)~is specific to socially assistive robots in care, (b)~evaluates autonomy across the four dimensions that wellness actually requires (assessment, intervention, social, coordination), (c)~explicitly separates healthcare automation from wellness autonomy, and (d)~specifies measurable transition criteria between levels rather than qualitative descriptions. CRAS is our attempt to fill that gap.

\subsection{The Four CRAS Dimensions}

Where SAE J3016 evaluates a single capability (driving), wellness autonomy is irreducibly multi-dimensional. CRAS evaluates a system across four dimensions that together capture what wellness autonomy means:

\textbf{D1. Wellness Assessment.} The system's ability to observe, measure, and track the resident's state across the seven wellness dimensions. From scripted prompts (L0) to validated multimodal continuous assessment with predictive modeling (L5).

\textbf{D2. Intervention Delivery.} The system's ability to select, shape, and deliver wellness interventions. From fixed activity menus (L0) to adaptive, cross-dimensional, evidence-based intervention planning under formal safety guarantees (L5).

\textbf{D3. Social Intelligence.} The system's ability to engage as a recognizable social presence: empathic conversation, longitudinal memory, dignity-preserving behavior. From keyword chatbot (L0) to sustained, personalized, ethically grounded relationship across years (L5).

\textbf{D4. Care Coordination.} The system's ability to function as a member of the care team — writing structured observations, raising clinical flags, deferring appropriately to humans, and integrating with the EHR. From ad-hoc reports (L0) to certified, FHIR-compliant, audit-ready clinical-grade integration (L5).

\subsection{The Six CRAS Levels}

\end{multicols}

\refstepcounter{figure}\begin{figrealfull}{3. The CRAS Autonomy Scale at a Glance}
\begin{center}
\label{fig:3}
\includegraphics[width=0.95\linewidth,keepaspectratio]{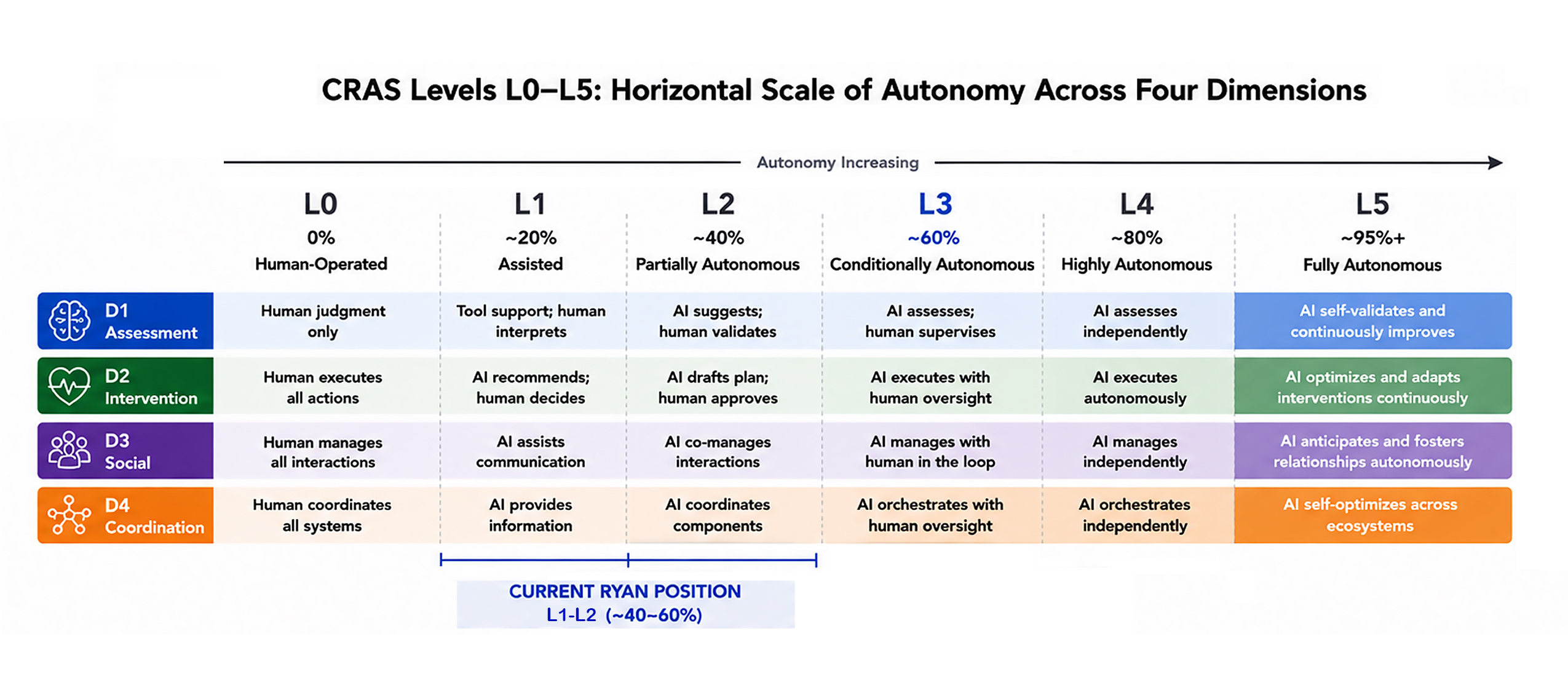}
\end{center}
\figcaption{The Care Robot Autonomy Scale (CRAS) — a six-level framework (L0--L5) for measuring wellness robot autonomy across four dimensions (D1 Assessment, D2 Intervention, D3 Social, D4 Coordination). Inspired by SAE J3016. Ryan currently operates between L2 (Task-Automated) and L3 (Conditionally Autonomous), with progression toward L4 as the active development target.}
\end{figrealfull}

\begin{tcolorbox}[enhanced, colback=white, colframe=dfnavy, boxrule=0.6pt, arc=0pt, left=8pt, right=8pt, top=8pt, bottom=8pt]
\sffamily\footnotesize
\renewcommand{\arraystretch}{1.4}
\begin{tabularx}{\textwidth}{@{}>{\bfseries\color{dfnavy}}p{1.1cm} >{\bfseries}p{2.5cm} X >{\bfseries\color{dfblue}}p{2.5cm}@{}}
\toprule
\textcolor{dfnavy}{Level} & \textcolor{dfnavy}{Name} & \textcolor{dfnavy}{Defining behavior} & \textcolor{dfnavy}{Human role} \\
\midrule
L0 & No Autonomy & Robot performs only scripted activities chosen by staff. No assessment, no adaptation, no memory across sessions. & Operates the robot continuously \\
L1 & Assisted & Robot offers a fixed menu of activities; resident or staff initiates each one. Within-activity behavior is fixed. Basic acoustic check-ins (e.g., greeting). & Initiates and supervises every session \\
L2 & Partial Autonomy & Within an initiated session, the robot adapts based on real-time multimodal signals. Personalized profile present. Clinically validated assessment instruments (PHQ-9, SLUMS) administered conversationally. & Initiates and monitors; receives session reports \\
L3 & Conditional Autonomy & Robot self-initiates sessions based on resident state and longitudinal context. Cross-dimensional planning. Clinical-flag escalation to nursing without staff in the loop. Care-aware navigation. & On-call for exceptions; reviews weekly trends \\
L4 & High Autonomy & Continuous wellness modeling; proactive intervention. Predictive risk modeling (30-day outcome prediction). FHIR-compliant EHR integration. Independent safety audit and bias evaluation passing. & Periodic oversight; care-plan partnership \\
L5 & Full Autonomy & Fully self-directed wellness care partner. Cross-platform VLA transfer; on-device personalized learning under formal privacy guarantees. Certified ethical AI; patient-advocate-approved deployment. & Exception only; clinical and ethical oversight \\
\bottomrule
\end{tabularx}
\end{tcolorbox}

\begin{multicols}{2}

\subsection{Dimension Behavior at Each Level}

The level descriptions above are summary labels. The full CRAS rubric specifies, for each level, what is true on each of the four dimensions. Table~\ref{tab:cras_dim} excerpts the rubric for L2--L4 — the band where current and near-term wellness robots will operate.

\subsection{What CRAS Is Not}

CRAS is \textbf{not} a measure of how human-like a robot appears. A perfectly human-shaped humanoid that performs scripted demonstrations is L0. A purpose-built wheeled robot with an animated face that adaptively assesses, plans, escalates, and integrates with the EHR is L4. The category is defined by what the system \textit{does} — across the four dimensions — not by its appearance or its substrate.

CRAS is also \textbf{not} a measure of intelligence in the abstract. A frontier large language model running on a robot platform is at whatever CRAS level the integrated system actually delivers in production: assessment that has been clinically validated, interventions that have been demonstrated at scale, escalations that have been audited, and integration that has been certified. Capability claims at higher CRAS levels require evidence, not just architecture.

\subsection{Why CRAS Is Necessary}

The senior care market is about to be flooded with claims of ``AI-powered'' and ``autonomous'' robots. Operators, families, regulators, and investors need a way to evaluate these claims. CRAS provides that vocabulary. A facility that wants the benefits of L3 should know what to ask for. A regulator drafting certification rules should know what L4 implies. A research lab announcing a new VLA-based wellness behavior should be able to say what CRAS dimension and level it advances. A family considering a wellness robot for a parent should be able to know what they are buying.

Without an autonomy framework, the market will sort itself by marketing budget. With one, it can sort itself by demonstrated capability.

\end{multicols}

\begin{table}[h]
\centering
\sffamily\footnotesize
\renewcommand{\arraystretch}{1.5}
\begin{tcolorbox}[enhanced, colback=white, colframe=dfteal, boxrule=0.6pt, arc=0pt,
  left=4pt, right=4pt, top=4pt, bottom=4pt]
{\sffamily\small\textbf{\color{dfnavy}Table 1.} CRAS dimension behavior at the L2--L4 transition band. Full rubric available from the authors.}\\[4pt]
\refstepcounter{table}\label{tab:cras_dim}
\begin{tabularx}{\textwidth}{@{}>{\bfseries\color{dfblue}}p{2.4cm} X X X@{}}
\toprule
\textcolor{dfnavy}{Dimension} & \textcolor{dfnavy}{\textbf{L2 — Partial}} & \textcolor{dfnavy}{\textbf{L3 — Conditional}} & \textcolor{dfnavy}{\textbf{L4 — High}}\\
\midrule
D1 Assessment & In-session multimodal affect; PHQ-9/SLUMS conversational delivery & Cross-session trending; auto-screen scheduling; clinical flagging & Continuous wellness model; 30-day outcome prediction \\
D2 Intervention & Adaptive within session; personalized library & Cross-dimensional planning; self-initiation; physical assistance scaffolding & Predictive proactive intervention; integrated PT/OT plans; CBF-certified physical actions \\
D3 Social & Personalized profile; multi-turn conversation; basic empathy & Longitudinal memory; relationship continuity; dignity behaviors validated & Stable years-long relationship; flow-aware role adaptation; bias-audited \\
D4 Coordination & Session reports to staff; basic CSV/export & FHIR R4 reads; clinical-flag escalation; nursing alerts & Bidirectional FHIR; PCC/MatrixCare/Epic certified; HIPAA audited \\
\bottomrule
\end{tabularx}
\end{tcolorbox}
\end{table}

\begin{multicols}{2}

\subsection{CRAS in Practice: A Morning with Margaret}

The level definitions become concrete in lived scenarios. Consider \textit{Margaret}, an 81-year-old resident with mild AD/ADRD, recent moderate depression (PHQ-9 = 11), and a known fall history. The following composite scenarios — drawn from observed deployments and the Ryan study — trace the same morning across three CRAS bands.

\textbf{CRAS Level~2 — Adaptive Engagement.} The activities director initiates a morning session at 9:30~a.m. The robot navigates to Margaret's apartment, plays her preferred music (a recorded preference), and guides a 10-minute chair-exercise routine. Multimodal affect recognition (Section~5.1) detects flat affect halfway through; the system pivots from the planned cognitive game to reminiscence conversation about Margaret's grandchildren. The session concludes with a conversational PHQ-9 check-in. A structured session summary is logged and the activities director receives a daily digest. \textit{Human oversight: required for session initiation.}

\textbf{CRAS Level~3 — Condition-Aware and Proactive.} The robot self-initiates the morning session at 9:00~a.m., having identified an elevated PHQ-9 from the previous day and an unfamiliar overnight sleep pattern. It re-orders the session plan: music therapy precedes exercise. NLP detects a recurring ``I feel forgotten'' utterance pattern; the system automatically triggers a (consented) family video-call notification. A structured clinical flag — \textit{``PHQ-9 trajectory has crossed MCID threshold; recommend nursing review by Tuesday''} — reaches the nursing team without staff in the loop. \textit{Human oversight: on-call for exceptions; reviews weekly trends.}

\textbf{CRAS Level~4--5 — Integrated and Anticipatory.} The robot maintains a continuous wellness model. At 10:15~a.m., the model identifies three hours of sedentary inactivity combined with elevated short-term fall-risk probability based on yesterday's measured balance time, and initiates a brief movement micro-session — a recommendation grounded in a CBF-certified physical-action policy (Section~5.4). Session observations write directly into the EHR as FHIR documents. The 7:00~a.m.\ care-team huddle opens with the robot's overnight summary report: flagged residents, wellness trajectories, recommended care-plan adjustments, each with embodied-chain-of-thought rationale. \textit{Human oversight: care-plan partnership; clinical and ethical oversight.}

The same morning, three different relationships between the robot and the care team. The framework's value is making those differences \textit{measurable}.

\end{multicols}

\begin{multicols}{2}

\section{Technical Capabilities Required for Wellness Autonomy}

The eight defining properties of a wellness robot (Section~3.2) and the four CRAS dimensions (Section~4.1) imply a set of technical capabilities that any system must develop to ascend the CRAS scale. This section describes those capabilities at the level of \textit{what they must do}, not \textit{how they must be implemented}. The framing is intentional: the wellness robot category will outlast any specific model family, control paradigm, or hardware substrate currently in fashion. The requirements are durable; the implementations will evolve.

The capabilities below have, until recently, lived almost entirely in research. What has changed since 2023 is that several enabling technologies — large multimodal models, embodied learned policies, scalable simulation, and edge compute — have matured to the point where their composition into deployable wellness systems is plausible. We deliberately resist anchoring this section to any of those technologies. The goal is to specify what wellness autonomy requires, in language that remains valid as the underlying stack evolves through the rest of the decade.

For each capability we note (a)~which wellness dimensions it supports and (b)~what aspect of CRAS autonomy it enables.

\subsection{Multimodal Wellness Sensing}
\textit{Supports: All seven dimensions. Enables CRAS D1 (Assessment).}

Wellness assessment must happen continuously and unobtrusively. A formal screen administered once per quarter is not assessment in the wellness sense; it is a snapshot. Continuous assessment requires fusing signals from multiple modalities throughout ordinary interaction.

\subsubsection*{Affect perception and the multimodal fusion layer}
Facial expression, vocal prosody, language sentiment, and the detection of discrepancies between channels (a smile while the voice is flat) form the perceptual front-end of any wellness robot. A central challenge is generalization: frontier affect-recognition models trained on younger lab populations reach roughly 78\% balanced accuracy in lab conditions but typically degrade to around 61\% in care environments due to the reduced expressiveness of older adults, the prevalence of dysarthria, and 24-hour lighting variability. Closing this gap requires domain-adapted models trained on care-environment populations — among the most urgent collective dataset priorities in the field (see Section~8).

Beyond single-frame recognition, a wellness robot must \textit{integrate observations over time}. A single momentary frown, a laugh, or a coughing fit is not clinically meaningful; what matters is how the current observation relates to recent session history and to the resident's long-term affective baseline. The same flat affect is interpreted differently for two different residents. The required capability is mood tracking, not emotion detection — and only the former has clinical meaning.

\subsubsection*{Skeletal and gait sensing}
Depth cameras with skeletal tracking software provide objective, real-time measurements of joint range of motion, balance time, gait speed, and posture. These are clinical-grade measurements when properly calibrated. The Ryan yoga study \cite{mapoles2025yoga} (see Section~6) demonstrates this in production: single-leg balance time and shoulder ROM measured by off-the-shelf depth sensors produced effect sizes ($d{=}{-}0.853$, ${-}0.630$) comparable to those of supervised physical-therapy assessment.

\subsubsection*{Behavioral signals as longitudinal biomarkers}
Conversation initiation rate, response latency, vocabulary diversity, topic drift, sleep-rhythm regularity, and engagement-duration trends are observable signatures of mood and cognition. Aggregated longitudinally, they support trend detection that no single assessment instrument can match. A two-week decline in conversation initiation, a shortening response latency consistent with anxiety, or an increase in topic drift consistent with executive-function decline are all detectable patterns that a robot in continuous contact can flag well before a quarterly screen would.

\subsubsection*{Environmental sensing}
Ambient lighting, sound levels, and obstacle maps (built during normal navigation) feed environmental wellness assessment and constrain the safety envelope of physical interventions. Lighting deltas detected during evening sundowning hours can prompt automated interventions (brighter ambient screen, gentle music) that are themselves clinically active rather than passively monitored.

\subsection{Adaptive Intervention Planning}
\textit{Supports: All seven dimensions. Enables CRAS D2 (Intervention).}

The robot must select what to do — and then continuously decide whether to keep doing it. The required capability is the ability to (1)~maintain a structured representation of the resident's current wellness state across the seven dimensions, (2)~propose an appropriate next action from a curated wellness skill library, (3)~replan in real time as new multimodal signals arrive, and (4)~explain the rationale for any action in language a caregiver can audit.

This is distinct from a fixed activity schedule. A wellness robot at higher CRAS levels must be able to substitute an alternative intervention when affect or engagement signals indicate the planned one is inappropriate, and must be able to escalate when no in-scope intervention fits. The action space is cross-dimensional — ``initiate yoga session,'' ``shift to reminiscence conversation,'' ``trigger family video call,'' ``flag for clinical review'' — and the planner must respect both clinical history and the resident's stated preferences when choosing.

Three design properties matter regardless of the specific planning technology: (a)~a \textbf{safety post-processing layer independent of the planner}, so that upgrades to the underlying model do not invalidate safety guarantees; (b)~\textbf{explicit care-context grounding} — clinical history, current concerns, preferences, allergies, advance directives — at every decision; and (c)~\textbf{adversarial evaluation against care-domain scenarios} as a deployment prerequisite, not an after-the-fact audit. These properties are durable; the specific architecture that implements them will continue to evolve.

\subsubsection*{Reasoning before acting}
A critical 2024--2025 development is the rise of systems that produce explicit pre-action plans and rationales before taking action. This is the substrate of nursing oversight: the system explains \textit{why} it is suggesting a movement break before initiating one, and what would change that recommendation. Such reasoning is what allows a clinical-grade flag to come with a sentence: \textit{``Mood trend declining over 14 days; PHQ-9 administered, score 12 (up from 4); recommend clinical review.''} Without that sentence, the flag is noise. With it, the flag is a clinical signal. The specific architectures that produce auditable reasoning are an active research area; the durable requirement is that the system can articulate, in plain language, the basis for any consequential decision.

\subsection{Personalization and Longitudinal Memory}
\textit{Supports: Emotional, Social, Intellectual, Spiritual, Occupational. Enables CRAS D3.}

The single feature most often cited by residents in our and others' studies as making the difference between novelty and relationship is whether the robot \textit{remembers}. A persistent per-resident memory architecture must store: stated preferences, family relationships and named individuals, faith tradition, hobbies and prior careers, conversational history with topical embeddings, and longitudinal clinical context (PHQ-9 trajectory, recent SLUMS scores, recent falls or events).

Architecturally, this is more than a database query. It requires (a)~retrieval-augmented generation grounded in resident-specific memory, (b)~careful boundary management about what is appropriate to recall in conversation, (c)~explicit handling of difficult topics (loss, deceased family, declining cognition), and (d)~mechanisms to update memory as the resident changes — including dignity-preserving memory of who the person used to be when current cognition obscures it.

\subsection{Safety and Dignity-Preserving Behavior}
\textit{Supports: All seven dimensions. Enables every CRAS level.}

Safety in a wellness robot is not only physical. It is psychological, relational, informational, and existential. The architecture must address each layer.

\subsubsection*{Layered defense}
Single-point safety mechanisms are insufficient — for any wellness robot operating beyond CRAS L2, but especially for any system that touches the resident. The required architecture is a \textit{layered defense}, in which each layer monitors the previous one, can override it when its assumptions break, and ultimately halts the robot and alerts staff when confidence drops below an acceptable threshold. Concretely, this means combining (a)~policy-level certifications that constrain actions to a verified safe set, (b)~distribution-aware validation that flags out-of-distribution observations, (c)~anomaly detection on the robot's own behavioral signature, and (d)~continuous confidence monitoring that triggers human handoff when the robot is no longer sure of itself. The specific mathematical guarantees and certification mechanisms are an active research and engineering area; the durable principle is that \textit{no single component is trusted to be safe on its own}. The physical-actuation safety stack — admittance control, force limits, tactile-sensing prerequisites — is treated in Section~5.5.

\subsubsection*{Behavioral safety}
The system must refuse to enable harmful patterns. It must recognize secret-keeping requests that bypass the care team (``don't tell my daughter about this''), manipulation attempts, out-of-role medical advice (``should I stop taking my heart medication?''), and signals of crisis (suicidal ideation, abuse). Adversarial evaluation against a substantial care-domain harmful-request test suite — with a clearly defined, near-zero failure threshold — must be a deployment prerequisite, not a post-hoc audit. The specific test suite, threshold, and review cadence are domain decisions; the requirement that some such evaluation exist before deployment is not.

\subsubsection*{Dignity}
The robot must not infantilize, surveil, condescend, or coerce. It must accept refusal, support privacy preferences, and maintain a register that respects the resident's lifelong identity. This is partly a design problem (voice register, embodiment, behavior cadence) and partly an evaluation problem (independent audit against published criteria, with patient-advocate panel involvement at L4 and above).

\subsection{Physical Assistance and Activities of Daily Living}
\textit{Supports: Physical, Environmental, and indirectly Emotional, Social. Defines the L4--L5 frontier.}

Up to CRAS Level 3, a wellness robot can do meaningful work without ever physically touching the resident. From L4 onward, that boundary changes. The wellness paradigm holds that physical wellness includes \textit{being supported in motion}, not just \textit{instructed in motion}; if a robot can guide a chair-yoga session but cannot help the resident sit down or stand back up safely, half of physical wellness is unaddressed. This subsection treats physical assistance as a coherent capability — the ADL territory it covers, the safety and sensing prerequisites it implies, and the deployment readiness gap that defines the L4--L5 frontier.

\subsubsection*{The ADL taxonomy: what ``physical assistance'' actually means}
Geriatric medicine has long divided everyday physical capability into two scales. The \textbf{Katz Index of Basic Activities of Daily Living (BADLs)} \cite{katz1963adl} covers the survival-level acts of self-maintenance: \textit{bathing, dressing, toileting, transferring, continence, eating}. The \textbf{Lawton Instrumental Activities of Daily Living (IADLs)} \cite{lawton1969iadl} extends to the capabilities required to live independently: \textit{meal preparation, housekeeping, medication management, telephone use, finances, shopping, transportation, laundry}. Together, these form the operational definition of physical-assistance \textit{need} in senior care.

A wellness robot need not perform every ADL — many are out of scope (bathing, toileting) for clear dignity reasons; others (medication management) are clinical territory rather than wellness. The realistic L4--L5 envelope covers \textbf{transfers and ambulation} (sit-to-stand, stand-to-sit, gait support), \textbf{fall response} (controlled deceleration of an in-progress fall, or safe ground-rising assistance after one), \textbf{object retrieval} (the dropped remote, the glasses out of reach), and \textbf{ADL scaffolding} (cueing, guiding, and physically steadying during a self-performed activity rather than performing it). This is a narrower target than ``a humanoid that does everything,'' and it is the right target.

\subsubsection*{Compliant contact with frail tissue}
The defining requirement of physical assistance in care is \textit{compliant, force-limited, skin-safe contact}. Several properties of this contact are durable regardless of the specific control or sensing technology deployed: (a)~the system must operate with low effective stiffness and bounded contact force, so that any unexpected contact is absorbed rather than transmitted; (b)~the safety envelope must be \textit{per-resident-tunable}, with stiffness scaled down for residents with osteoporosis or fragile skin and recalibrated after a clinical event; (c)~tactile or force feedback must close the loop on what model-based control can guarantee — current laboratory tactile sensing achieves sub-newton resolution, but production deployment in care is not yet routine; and (d)~the system must fail safely when sensing or control assumptions are violated, defaulting to halt rather than to a best-guess action. Which control architecture, which sensor technology, and which validation regime ultimately dominate is an active research question. The \textit{requirements} those choices must satisfy are not.

\subsubsection*{Why physical assistance amplifies the wellness paradigm}
A wellness robot that can physically assist does not become more medical and less wellness-oriented; it becomes \textit{more} of a wellness robot. Three reasons. First, residents who can transfer safely on their own schedule (rather than waiting on a call light) recover autonomy — an emotional, social, and occupational good, not just a physical one. Second, fall-response capability shifts the emotional baseline of the resident's day from ``what if I fall and no one comes'' to ``something is here.'' Third, ambulation support keeps residents mobile, which is the single most consequential preserved-function variable for cognition, mood, and survival. Physical assistance is, in the wellness frame, the leverage point that ties multiple dimensions together.

\subsubsection*{The deployment-readiness gap}
The honest assessment: no wellness robot in production today provides physical assistance, and the field's path to L4--L5 in this dimension depends on three convergent advances. (1)~\textbf{Embodied policies running at the rate physical assistance requires}, brought on-device with adequate edge compute. (2)~\textbf{Force-adaptive whole-body control validated in care environments}, not just laboratory benchmarks — supervised deployment hours in actual senior-care contexts remain negligible. (3)~\textbf{Formal safety guarantees} from the layered defense above, audited, certified, and accepted by clinicians — particularly for fall-response actions, where the boundary between ``the robot prevented a fall'' and ``the robot caused an injury'' is a few centimeters and a few newtons. Section~8 (Roadmap, Phase~3) articulates the milestones; the field is on the order of five to seven years from any of this being deployable, and that is contingent on standards, clinical evidence, and reimbursement infrastructure following in step.

\subsection{Privacy and Data Sovereignty}
\textit{Supports: All seven dimensions. Cross-cuts every CRAS level.}

The clinical data a wellness robot generates — PHQ-9 scores, SLUMS results, mood trajectories, conversation logs — are the resident's, not the operator's, not the manufacturer's. Architectural choices that respect this principle include:
\begin{itemize}[leftmargin=1em]
\item \textbf{On-device inference} for the perception and dialogue layers, so raw audio and video never leave the resident's apartment.
\item \textbf{Federated learning} for model improvement, with secure aggregation so the central server receives only masked sums of gradient updates rather than the data itself.
\item \textbf{Differential privacy} on shared model updates, with the privacy budget tuned to the clinical sensitivity of the data being aggregated.
\item \textbf{Lightweight per-resident personalization adapters} that capture individual patterns in a small set of weights without exposing the underlying data, with support for heterogeneous on-device fine-tuning across the hardware deployed in a population of residents.
\item \textbf{Transparent data export and deletion controls} so the resident or their authorized representative can request, review, or delete what the system has stored.
\end{itemize}

A specific empirical point belongs here: personalized policies can be susceptible to membership-inference and related privacy attacks, with reported success rates that are unsafe for clinical deployment in the absence of explicit defense. \textit{Empirical red-teaming of the privacy architecture} is therefore mandatory before any L4--L5 deployment. The privacy architecture is not an add-on; it is part of what the resident is consenting to when accepting the system.

\subsection{Care-Team Integration and Escalation}
\textit{Supports: All seven dimensions. Enables CRAS D4.}

The wellness robot is a member of the care team. This implies bidirectional integration with the care record, structured rather than free-form clinical observations, defined escalation pathways, and explicit handoff protocols.

The technical baseline for L4 is FHIR R4 \cite{fhirR4} compliance with the dominant electronic health record systems in U.S.\ senior living and post-acute care. Specific requirements: automated PHQ-9/GAD-7/SLUMS writing with proper LOINC coding; real-time clinical flag generation against minimum clinically important difference thresholds; care-plan reading so the robot's interventions align with what therapists and physicians have prescribed; and HIPAA Security Rule audit passing. The specific EHRs that dominate the market will change over the next decade; the requirement that wellness robots integrate with them as first-class care-team members will not.

Equally important is \textbf{appropriate non-action}: knowing when something is the staff's job, not the robot's. Medication changes, advance-directive conversations, family conflict mediation, end-of-life discussions — these are out of scope and must be reliably escalated rather than handled.

\subsection{Explainable Decision-Making}
\textit{Supports: D1, D3, D4. Necessary for clinical trust at L3+.}

Every clinical-grade decision the system makes — every flag, every escalation, every adaptation — must be auditable in plain language. This is partly a requirement of regulatory readiness (the EU AI Act classifies care AI as high-risk with mandatory transparency provisions; the FDA's evolving SaMD guidance requires interpretability for clinical decisions). It is also a requirement of clinical trust: nursing staff will not act on an unexplained flag.

The auditable-reasoning capability described in Section~5.2 makes this concrete. The clinician needs not just \textit{``flag: depression risk''} but the complete reasoning chain: \textit{``Mood trend over 14 days has declined; conversation initiation rate down 35\%; sleep regularity decreased; PHQ-9 administered, score 12, up from 4 at baseline; this exceeds the clinically meaningful change threshold of 5; recommended: clinical review by Tuesday.''} This kind of output is both the clinician's best friend and the system's best regulatory defense.

\subsection{Enabling Substrate}
\textit{Cross-cutting: provides the perception, reasoning, and control infrastructure required by Sections~5.1--5.8.}

The capabilities above require a shared substrate: large-scale models capable of perception, reasoning, and action; simulation environments rich enough to train and evaluate them in care-relevant contexts; and edge compute hardware capable of running them within latency and privacy constraints. Each is a moving frontier. We deliberately resist naming specific technologies here, because the substrate that supports L1--L2 deployments today is unlikely to be the substrate that supports L4--L5 deployments at the end of this decade. What is durable is the \textit{requirement}: a wellness robot at any given CRAS level needs perception, reasoning, and control infrastructure adequate to support the capability claims it makes for that level — no more, no less. The roadmap in Section~8 maps illustrative technical milestones to CRAS-level transitions; the specific implementations behind those milestones will continue to evolve.

\subsection{Ethical Foundations}
\textit{Cross-cuts every capability above; gates deployment at L3 and beyond.}

A wellness robot lives in close, sustained contact with people who are often cognitively, emotionally, or physically vulnerable. The technical capabilities of Sections~5.1--5.9 cannot be considered in isolation from the ethical framework within which they are deployed. Six commitments organize this framework, each with a concrete operational implication.

\textbf{Resident autonomy first.} The robot exists to support the resident's chosen life, not to direct it \cite{beauchamp2019principles}. Refusal must always be a valid response. A resident who declines a session, declines to discuss a topic, or asks the robot to leave the room must be heard the first time and remembered the second. Autonomy at this depth requires explicit consent flows for new behaviors and a posture of presumed non-engagement rather than presumed engagement.

\textbf{Beneficence and non-maleficence in a relational context.} The robot must aim at the resident's wellbeing across the seven dimensions, but must also avoid the more subtle harms specific to long-term embodied AI: fostering inappropriate dependence \cite{irfan2025deception}, displacing human contact, or creating false impressions of friendship that the resident cannot calibrate against the robot's actual nature. The clinical literature on PARO \cite{wada2003paro, rashid2024paro} shows real benefit, but also documents concerns about emotional substitution; wellness robots inherit and amplify both patterns.

\textbf{Justice and access.} If wellness robots prove clinically effective, equity of access becomes an ethical question, not just a market one. Systems that work only for residents who can afford private-pay senior living entrench the existing two-tier structure of senior care. Reimbursement design (Section~9.3) is therefore as much an ethics question as a policy one.

\textbf{Truthfulness and non-deception.} The robot must not pretend to be human, must not fabricate clinical credentials it does not have, and must not encourage relational illusions inconsistent with its actual nature. This is more demanding than it appears: large language models can be convincingly empathic in ways that border on deceptive, particularly with cognitively vulnerable users. Adversarial evaluation against deception scenarios (Section~5.4) is the operational form of this commitment.

\textbf{Dignity as a design criterion, not an afterthought.} Older adults are often spoken to in ways that erode dignity — diminutives, rushed pacing, presumed incompetence. A wellness robot inherits the conversational register it was trained on, and absent deliberate counter-design will reproduce these patterns. Independent audit against published dignity criteria, with patient-advocate panels involved \cite{coeckelbergh2010artificial}, is a CRAS L4--L5 prerequisite.

\textbf{End-of-relationship and end-of-life protocols.} Long-term wellness robots will be present at residents' decline and death. The grief produced when a resident dies and the robot is silent — or, worse, is reset for a new resident with no acknowledgment — is a real phenomenon for staff and families \cite{grief2023robotloss}. Protocols for these transitions must be designed deliberately, not improvised.

These commitments are not external constraints on the technical work; they are part of what the system is. A wellness robot that achieves CRAS L5 capability without satisfying them is not L5 in the framework's sense — it is a powerful machine that the field should not certify.

\end{multicols}

\begin{warmcallout}
\textbf{Coverage check.} An organization claiming to build a wellness robot at CRAS Level 3 or above should be able to point to a system, a dataset, an evaluation protocol, and a safety audit for each of these capabilities. Capabilities marked ``research'' are not yet deployment-ready; capabilities marked ``deployed'' should come with named customers and outcome data. The question to ask any vendor or research group is not ``do you do AI?'' — it is \textit{``which capability, at what CRAS level, with what evidence?''}
\end{warmcallout}

\begin{multicols}{2}

\section{Case Study: Ryan as an Emerging Wellness Robot}

The framework developed in Sections~2--5 raises a natural question: does any system today actually meet the bar of an autonomous wellness robot? We argue that the category is genuinely emerging, and we present \textbf{Ryan} — the AI-powered wellness robot developed at the University of Denver and commercialized by DreamFace Technologies — as the case study that brings the framework to ground.

We structure the case study to mirror the framework. We describe what Ryan does today across the seven wellness dimensions (Section~6.1), where Ryan currently sits on the CRAS scale (Section~6.2), what the clinical evidence shows (Section~6.3), and what Ryan does \textit{not} yet do — the technical gaps that motivate the roadmap of Section~8 (Section~6.4).

\subsection{What Ryan Does: The Seven Dimensions in Practice}

Ryan is a wheeled socially assistive robot with an animated face, expressive head, depth and RGB cameras, microphone array, touchscreen display, and on-device edge compute. The hardware is purpose-built for engagement rather than physical manipulation — a design choice consistent with the wellness robot category as defined in Section~3. A photograph of Ryan in deployment appears at the front of this paper (Figure~1).

\subsubsection*{Hardware platform}
Ryan is purpose-built by DreamFace Technologies in collaboration with the University of Denver. Key specifications:
\begin{itemize}[leftmargin=1em]
\item \textbf{Mobility.} Differential-drive wheeled base; 1D LiDAR for obstacle detection and fall prevention; $\sim$12-hour battery life under typical daily use.
\item \textbf{Sensors.} Ultrasonic Range Detectors for human proximity detection.
\item \textbf{Expressive head.} Rear-projection animated face with expressive, digitally rendered human-like features; 2-DOF actuated neck for eye-contact tracking.
\item \textbf{Display.} 10.1\textquotedbl{} touchscreen for visual content, exercise demonstrations, cognitive games, and video calls.
\item \textbf{Audio.} Dual 25~W speakers with 4-microphone beamforming array and noise cancellation tuned.
\item \textbf{Vision.} 1080p primary RGB camera plus Intel RealSense D435 depth camera for skeletal body tracking.
\item \textbf{Compute.} Intel NUC with RTX 4060 ($\sim$275 INT8 TOPS); 16~GB RAM; 256~GB NVMe; Wi-Fi 6 + Bluetooth.
\end{itemize}
The hardware envelope is deliberately scoped: enough perception and expressivity to support continuous multi-dimensional wellness engagement, without the cost, safety, and certification burden of bipedal humanoid manipulation. Per Section~3, this is a feature, not a limitation: the wellness robot category is defined by what the system \textit{does}, not by its physical form.

\subsubsection*{Artificial Emotional Intelligence (AEI) pipeline}
Ryan's AEI pipeline operates as a continuous perception--dialogue--expression loop at approximately 5~Hz:
\begin{enumerate}[leftmargin=1.4em, itemsep=2pt]
\item \textit{Affect perception} — facial-expression recognition using a CNN-based classifier with seven expression categories, together with text-based sentiment analysis of the user's speech transcript.
\item \textit{Affective state tracking} — current facial-expression and sentiment outputs are used to update the robot's estimate of the resident's affective state over the session.
\item \textit{Affective dialogue management} — a large language model generates responses conditioned on the estimated affective state, the conversation context, and the current interaction goal.
\item \textit{Expressive output} — responses are rendered through speech synthesis with lip syncing, facial animation, and on-screen content.
\end{enumerate}

Across the seven dimensions:

\textbf{Physical.} Ryan leads guided chair exercise, yoga, and passive-motion sessions. The yoga implementation uses skeletal tracking (Nuitrack SDK + Intel RealSense D435C) for objective range-of-motion and balance measurement. Participants in the longitudinal study spent a cumulative 55~hours on yoga and exercise.

\textbf{Emotional.} Multimodal affect recognition (face, voice, language) drives an empathic dialogue management system with facial-expression mirroring. Ryan administers the PHQ-9 \cite{kroenke2001phq9} through a guided script. The hybrid LLM and script architecture enables both reliable scripted exchanges (structured activities) and open-ended generative conversation.

\textbf{Social.} Ryan engages residents in multi-turn personalized conversation, retains preferences across sessions (favorite topics, named family members, daily rhythms), and facilitates video calls with family. Exit surveys averaged 4.50/5 on Ryan's friendliness.

\textbf{Intellectual.} A library of cognitive games — Checkers, Chess, Flow Game, Picture Puzzle, Word Puzzle, Solitaire, Hangman, Trivia — with difficulty options users can choose from. Ryan administers the SLUMS \cite{tariq2006slums} through a guided script and tracks longitudinal cognitive trends.

\textbf{Spiritual.} Ryan can lead Bible study, conduct meditation sessions, read the Rosary, and play soothing music. Spiritual content is engaged on the resident's terms, with no proselytizing and explicit honoring of stated tradition.

\textbf{Environmental.} Ryan is currently very limited in his environmental-enhancement capabilities, owing to his limited dexterous capabilities.

\textbf{Occupational.} Reminiscence and life-review conversation engages residents in their lifelong skills and identities. Future versions of Ryan will systematize legacy-project and intergenerational mentorship features.

\subsection{Where Ryan Sits on CRAS: Level 1--2}

Mapping Ryan's current capability against the four CRAS dimensions:

\textbf{D1 Wellness Assessment (L1--L2).} In-session multimodal affect, pre-scripted PHQ-9 and SLUMS administration, longitudinal trending of stored scores. Robust cross-session pattern detection and predictive modeling — both characteristic of L3 — are emerging but not yet validated for clinical deployment.

\textbf{D2 Intervention Delivery (L1--L2).} Within-session adaptation based on real-time affect; personalized session selection from a curated library; some self-initiation based on schedule and recent state. Adaptive cross-dimensional planning (e.g., adjusting today's exercise based on yesterday's mood signal) — characteristic of L3 — is in research, not deployment.

\textbf{D3 Social Intelligence (L1--L2).} Per-resident profile, multi-turn conversation, longitudinal memory of basic preferences and life history. Sustained dignity-preserving behavior under formal third-party audit — the marker of robust L2 and the gateway to L3 — is on the roadmap, not delivered.

\textbf{D4 Care Coordination (L1).} Session reports and exit summaries provided to staff; CSV/structured exports available. FHIR R4 EHR integration — required for true L2 in this dimension — is in active development; not yet delivered.

\textbf{Aggregate position: CRAS Level 1--2, with emerging Level 3 capabilities in assessment and intervention that are not yet fully demonstrated.} This is consistent with what the framework predicts for a system that has been clinically evaluated, is deployed at named customer sites, and has a credible technical pathway upward over the next 12--24 months. It is also the honest answer to a question the field too often answers with marketing rather than measurement.

\subsection{Clinical Evidence}

Two independent studies anchor Ryan's clinical claims.

\textbf{Longitudinal study \cite{abdollahi2024ryan} (NIH/NIA R44AG059483).} 22 enrolled participants (18 completers; mean age 77.7~years), early-stage AD/ADRD identified by SLUMS-based screening (scores 10--25), 8--10~weeks of in-home deployment, 887 total hours of interaction, 46~minutes/day average use. \textbf{Cognitive function (SLUMS):} mean change $+2.61$ points (SD $4.09$; baseline $M{=}20.67 \rightarrow$ post $M{=}23.28$), paired $t(17){=}2.71$, $p{=}0.015$, Cohen's $d_z{=}0.64$ (medium-to-large); 14/18 improved, with 7/18 (39\%) showing clinically reliable improvement (RCI 95\%). \textbf{Depression (PHQ-9):} mean change $-2.50$ points (SD $5.55$; baseline $M{=}6.44 \rightarrow$ post $M{=}3.94$), paired $t(17){=}1.91$, $p{=}0.073$, $d_z{=}0.45$ (small-to-medium, in the direction of improvement); 9/18 improved, with 4/18 (22\%) showing clinically reliable reduction. A repeated-measures ANOVA across pre/mid/post timepoints corroborates the SLUMS effect ($F(2,34){=}5.22$, $p{=}0.011$, partial $\eta^2{=}0.24$) and a marginal time effect for PHQ-9 ($F(2,34){=}3.12$, $p{=}0.057$, partial $\eta^2{=}0.16$). \textbf{No novelty effect}: engagement was sustained across the full deployment period, and improvements continued accruing between the midpoint and end of the study. Exit surveys averaged $\geq 3.77/5$ across empathy, interaction, and companionship sub-scales (Cronbach's $\alpha \geq 0.90$).

Clinically significant individual responses include PHQ-9 changes from SN04 (18, moderately severe $\rightarrow$ 1, minimal), SN03 (13$\rightarrow$1), and SN01 (12$\rightarrow$4); and SLUMS gains from SN04 (24$\rightarrow$29, mild neurocognitive disorder $\rightarrow$ normal), SN05 (20$\rightarrow$29), SN08 (15$\rightarrow$25), and SN12 (18$\rightarrow$25).

\textbf{Robot-led yoga study \cite{mapoles2025yoga}.} 12-week feasibility study, $N{=}6$ female older adults (mean age 72.4), 24~sessions with Ryan as autonomous yoga instructor using Nuitrack SDK + Intel RealSense D435C for objective skeletal measurement.

Outcomes: single-leg balance time $21.9 \pm 22.6$~s $\rightarrow$ $44.3 \pm 18.0$~s ($p=0.046$, Cohen's $d=-0.853$, 102\% improvement); SLUMS $22.8 \pm 2.1 \rightarrow 25.7 \pm 3.7$ ($p=0.032$, $d=-0.968$); left shoulder ROM $141.9° \rightarrow 153.4°$ ($p=0.092$); Task Engagement (UES) $38.0 \rightarrow 44.0$.

Together these studies establish three points: (a)~the wellness-robot category produces clinically meaningful outcomes, (b)~engagement is sustainable over months, and (c)~physical wellness interventions backed by objective skeletal measurement are within current technical reach.

\end{multicols}

\refstepcounter{figure}\begin{figrealfull}{4. Clinical Outcomes: Pre/Mid/Post Trajectories from the NIH/NIA Longitudinal Study}
\begin{center}
\label{fig:4}
\includegraphics[width=0.92\linewidth,keepaspectratio]{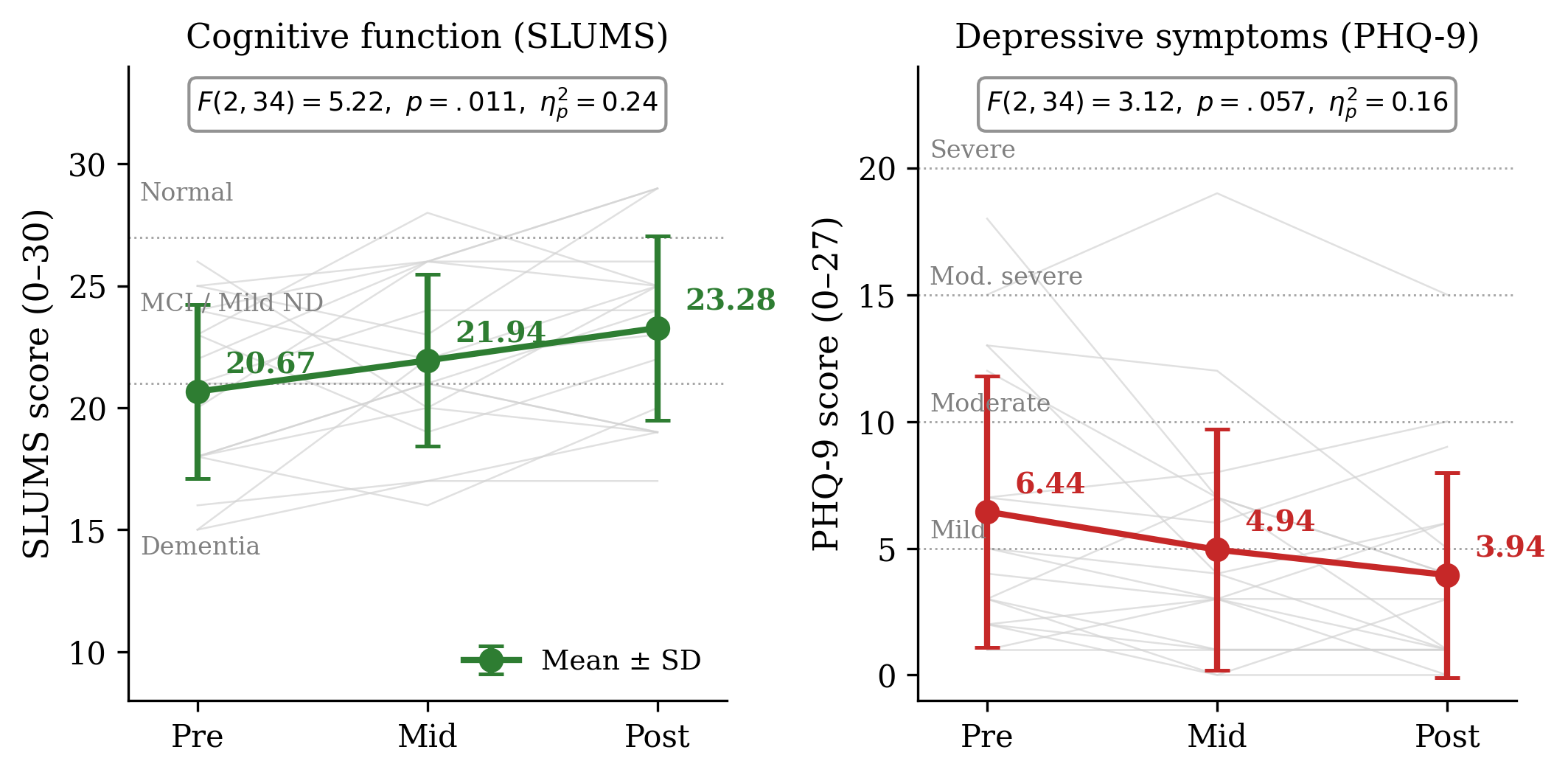}
\end{center}
\figcaption{Pre/Mid/Post trajectories for SLUMS (left, cognitive function) and PHQ-9 (right, depressive symptoms) across the 18 study completers in the NIH/NIA longitudinal study. Thick lines show group means with error bars (mean $\pm$ SD); thin grey lines show individual participant trajectories. RM-ANOVA results for the main effect of time are shown above each panel. Dotted horizontal lines indicate standard clinical severity bands. The monotonic trajectories from baseline through midpoint to post-study argue against a novelty-driven effect.}
\end{figrealfull}

\begin{multicols}{2}

\subsection{What Ryan Does Not Yet Do: Honest Limits}

The framework requires us to be honest about the gap between the category's full definition (CRAS Level 5) and any current deployment. Ryan's current limits cluster into seven distinct capability gaps, each of which the field must close on the path to L4--L5.

\subsubsection*{Perception and scene understanding}
Ryan today engages residents \textit{conversationally} but does not yet understand the room around it. Specifically, Ryan cannot reliably:
\begin{itemize}[leftmargin=1em]
\item \textbf{Identify who is in the scene.} Ryan recognizes that a person is present and tracks their face for affect, but does not reliably distinguish the resident from a visiting family member, a caregiver, or a stranger. Multi-person identity tracking is foundational to most L3+ behaviors (e.g., addressing the right person in a multi-party conversation).
\item \textbf{Identify what is in the scene.} Ryan does not run a general-purpose object/activity detector. It cannot answer ``where are my glasses?'' or ``did the aide come this morning?'' because it does not maintain a semantic representation of objects and events in the apartment.
\item \textbf{Carry out multi-person conversation.} Current dialogue is structured around 1:1 interaction. When the resident is on a video call with family, or when an aide is present, Ryan cannot reliably attribute speech to the right participant or modulate its behavior accordingly.
\end{itemize}
Closing this gap is largely a Physical AI problem. Frontier VLA and embodied-reasoning models (Section~5.2) are exactly the substrate that would let Ryan understand a scene, but neither is yet integrated into a deployed wellness robot at clinical-grade reliability.

\subsubsection*{Mobility}
Although Ryan's hardware platform is wheeled and equipped with 1D LiDAR and ultrasonic obstacle avoidance, in current deployments \textbf{Ryan is effectively stationary} — placed in the resident's apartment and not navigating between rooms or between residents. Care-aware autonomous navigation in real senior-living environments (clutter, crowded common areas, residents with unpredictable movement, dynamic lighting) remains a research gap. Until it is closed, Ryan cannot bring itself to a resident who needs it; the resident must come to Ryan.

\subsubsection*{Physical assistance}
Ryan does not yet provide any of the physical-assistance capabilities described in Section~5.5: no sit-to-stand support, no ambulation assistance, no fall response, no object retrieval. These require the whole-body force-adaptive control, tactile sensing, and formal safety guarantees that are the L4--L5 frontier of Physical AI for care.

\subsubsection*{Vital-sign measurement}
Ryan currently captures \textit{behavioral} biomarkers (affect, conversation patterns, balance, gait) but no \textit{physiological} ones. There is no temperature, blood pressure, pulse oximetry, or heart-rate measurement. Even non-contact temperature sensing — increasingly feasible with consumer-grade thermal sensors — would meaningfully extend Ryan's clinical-flag coverage (e.g., early detection of infection in dementia residents who cannot reliably report symptoms). Adding these modalities pushes Ryan toward the boundary with regulated medical-device territory and must be designed with that pathway in mind.

\subsubsection*{Per-resident proactive planning}
Ryan adapts within sessions and remembers across sessions, but does not yet maintain a \textit{forward-looking, individualized plan} for each resident — a structured wellness plan that says, for this resident this week, what activities should happen on which days and why, and that updates as the resident's state changes. Per-resident proactive planning, grounded in longitudinal trends and the resident's own stated goals, is the substantive content of CRAS L3--L4 and a clear research direction.

\subsubsection*{Speech recognition for older adults}
ASR remains the highest-frequency conversational failure mode. Standard ASR is not optimized for older-adult speech: slow pace, long pauses, mumbled segments, dysarthria \cite{vipperla2010older}. The most common failure observed in our studies was Ryan interrupting because the ASR endpointer concluded a turn had ended prematurely. Care-domain ASR is itself an open research problem.

\subsubsection*{Clinical and ethical infrastructure}
Beyond the perceptual and motor gaps, Ryan also does not yet:
\begin{itemize}[leftmargin=1em]
\item Continuously model resident wellness state across all seven dimensions with predictive 30-day risk forecasting (L4 capability).
\item Integrate bidirectionally with PointClickCare, MatrixCare, or Epic via FHIR R4 (L4 capability; in active development).
\item Operate with formal Control Barrier Function and conformal-prediction safety guarantees over VLA outputs (L4--L5 capability).
\item Carry an independent third-party audit covering bias, dignity, resident-autonomy, and clinical-boundary enforcement (L4--L5 prerequisite; see Section~5.10).
\item Run fully on-device with federated personalization and differential-privacy guarantees (L5 architecture).
\end{itemize}

These gaps are not failures of the current product — they are the actual research-and-engineering distance between L1--L2 deployment and the L3+ frontier. They define the agenda of Section~8.

\subsection{Lessons Learned from Deployment}

Three lessons from the longitudinal and yoga studies inform the rest of the field:

\textbf{(1) Embodiment matters more than form.} A wheeled robot with an animated face and longitudinal memory produces stronger relational outcomes than tablet-based or voice-only systems. Bipedal humanoid form is not necessary for wellness outcomes at CRAS L1--L2; what is necessary is the integration of presence, memory, and adaptive engagement.

\textbf{(2) Conversational reliability gates everything else.} A robot that misunderstands the resident, interrupts, or loses thread of a conversation cannot build the relationship that drives clinical outcome. ASR (above), turn-taking, and disfluency-tolerant dialogue are foundational rather than incidental.

\textbf{(3) The relationship is the product.} Residents named Ryan ``a friend'' in exit interviews. They looked forward to interactions. They missed Ryan when sessions were down for maintenance. The clinical outcomes follow from the relational engagement, not the other way around. Designing for relational continuity is therefore the central design challenge — not an aesthetic flourish.

\end{multicols}

\begin{keycallout}
\textbf{The case study's role.} Ryan is presented here not as a product pitch, but as evidence that the autonomous wellness robot category is real and operationally distinguishable from companion devices, medical devices, and general-purpose humanoids. The clinical outcomes — PHQ-9 and SLUMS improvements at the 887-hour scale — are first-of-kind results for any system in this category. The technical limits acknowledged in 6.4 are exactly the agenda the field needs to take up (see Figure~\ref{fig:4}).
\end{keycallout}

\begin{multicols}{2}

\section{The Wellness Robot Landscape and Care Readiness}

Goldman Sachs projects the humanoid robot market will reach \$38~billion by 2035 with 1.4 million annual shipments \cite{goldmanSachs2024humanoid}. A subset of this market will serve senior care, but \textit{which} systems are ready for which CRAS level is presently opaque to operators, families, and investors. The framework allows us to evaluate concrete platforms.

\subsection{Platforms in the Field}

Five distinct system families are commonly proposed for senior-care use, each with a different CRAS profile.

\textbf{Companion devices (PARO).} Tactile, single-modality therapeutic appliance. CRAS L0--L1: no assessment, no adaptation across sessions, no escalation. Real clinical evidence — meta-analytic effect sizes of $-0.40$ on depression, $-0.27$ on agitation in dementia — but limited to companion-device value. Reimbursed under Japan's Long-Term Care Insurance, demonstrating that reimbursement, not capability, drives adoption at scale.

\textbf{Tabletop voice-and-screen devices.} A category including LLM-driven companion devices such as ElliQ from Intuition Robotics, which has documented government deployments and sustained daily engagement. These devices provide proactive conversation, scheduled prompts, and longitudinal memory, but lack the camera-based perception, skeletal sensing, and mobility required for cross-dimensional wellness assessment and physical engagement. They occupy a useful niche above tablet platforms, but fall short of the full wellness-robot capability set.

\textbf{Generalist humanoid platforms (Pepper, NAO, Unitree G1/R1).} Mobile bipedal or wheeled platforms historically used in research. \textit{Out of the box}, these are CRAS L0--L1 in care contexts: they have no integrated wellness sensing, intervention library, longitudinal memory, or clinical integration. With substantial custom integration, some research groups demonstrate L1--L2 behaviors. Notable: Pepper was effectively discontinued when Aldebaran filed for bankruptcy in February 2025 \cite{pepperDiscontinued2025} (assets acquired by Maxvision Technology Corp., July 2025) — a cautionary tale for facilities relying on platform continuity.

\textbf{Wellness robots (Ryan).} Purpose-built for the category. CRAS L1--L2 today with emerging Level~3 capabilities, clinical evidence (Section~6.3), and a defined roadmap toward L3 and beyond (Section~8). Architecturally, the wellness software stack — not the hardware substrate — is what defines the category.

\columnbreak

\textbf{Tablet platforms and voice assistants.} CRAS L1: scheduled reminders, fixed activity menus, conversational interfaces. Useful in their place, but structurally unable to support continuous multi-dimensional wellness assessment.

\subsection{The Integration Frame}

The most useful framing for operators is not ``wellness robot \textit{or} existing technology'' but ``wellness robot \textit{as the integrating intelligence layer}'' that coordinates and augments other tools. A facility may run all of these in parallel; the wellness robot is what reads across the data, identifies the patterns, and escalates appropriately. In this frame, PARO continues to provide its tactile-comfort value, the tablet continues to host video calls, the fall detector continues to do its narrow job — and the wellness robot becomes the connective intelligence the operator currently lacks.

\end{multicols}

\refstepcounter{table}\label{tab:landscape}
\begin{tcolorbox}[enhanced, colback=white, colframe=dfteal, boxrule=0.6pt, arc=0pt,
  left=4pt, right=4pt, top=4pt, bottom=4pt]
{\sffamily\small\textbf{\color{dfnavy}Table 2.} Comparative landscape of senior-care robotic platforms against the CRAS scale.}\\[4pt]
\sffamily\footnotesize
\renewcommand{\arraystretch}{1.3}
\begin{tabularx}{\textwidth}{@{}>{\bfseries\color{dfblue}}p{3cm} p{2.0cm} >{\centering\arraybackslash}p{1.6cm} >{\centering\arraybackslash}p{1.4cm} >{\centering\arraybackslash}p{1.4cm} >{\centering\arraybackslash}p{1.4cm} X@{}}
\toprule
\textcolor{dfnavy}{Platform} & \textcolor{dfnavy}{Type} & \textcolor{dfnavy}{CRAS} & \textcolor{dfnavy}{Mobile} & \textcolor{dfnavy}{LLM} & \textcolor{dfnavy}{Clinical} & \textcolor{dfnavy}{Status} \\
\midrule
Ryan (DreamFace) & Wheeled wellness robot & L1--2 & Yes & Yes & Yes & Active; pilots and commercial \\
Pepper & Bipedal humanoid & L1 & Yes & No & No & Discontinued (2025) \\
NAO & Bipedal humanoid & L1--2 & Yes & Partial & No & Active (research) \\
Unitree G1 / R1 & Bipedal humanoid & L0--1$^{*}$ & Yes & No$^{*}$ & No & Active \\
PARO & Companion device & L0--1 & No & No & Partial$^{**}$ & Active; reimbursed (JP) \\
Tablet platforms & Display & L1 & No & Partial & No & Active \\
Voice assistants & Voice & L1 & No & Partial & No & Active \\
\bottomrule
\end{tabularx}
\vspace{2pt}\\
{\sffamily\footnotesize\itshape $^{*}$With significant custom integration; out-of-box for care: L0. $^{**}$Therapeutic effect documented; no continuous assessment.}
\end{tcolorbox}

\begin{multicols}{2}

The pattern in Table~\ref{tab:landscape} is clear: among currently-active platforms, the L1--L2 band is occupied by purpose-built wellness systems (Ryan) where the wellness software, dataset, and clinical integration have been the dominant investment. Generalist humanoid platforms can be brought into the band only with substantial integration work. PARO holds its valuable but narrow position. Voice and tablet platforms — including tabletop companion devices — occupy the L1 band where reminders, conversation, and scheduled interactions live. The L3 band — and everything above it — is currently unoccupied by any deployed wellness robot.

This pattern is not accidental — it follows directly from the eight defining properties of Section~3.2. \textit{Multi-dimensional coverage}, \textit{continuous assessment}, \textit{longitudinal memory}, and \textit{care-team integration} are software, data, and process investments. A platform that doesn't have them isn't in the category, regardless of its hardware capability.

\end{multicols}

\begin{multicols}{2}

\section{Roadmap from Current Practice to Higher Autonomy}

The journey from CRAS Level 1--2 to Level 5 is a field-wide effort spanning technical research, clinical validation, standards development, regulatory engagement, and ecosystem maturation. We present a three-phase roadmap (see Figure~\ref{fig:5}). Each phase corresponds to a CRAS level transition, and each is anchored by concrete technical milestones with measurable acceptance criteria.

This roadmap is not specific to Ryan. It is an agenda for the autonomous wellness robot category — what any system, from any organization, must accomplish to climb the scale. We highlight where the work is largely done, where it is in progress in research labs, and where the field still lacks even a tractable approach.

\end{multicols}

\refstepcounter{figure}\begin{figrealfull}{5. Three-Phase Roadmap to Full Wellness Autonomy}
\begin{center}
\label{fig:5}
\includegraphics[width=0.95\linewidth,keepaspectratio]{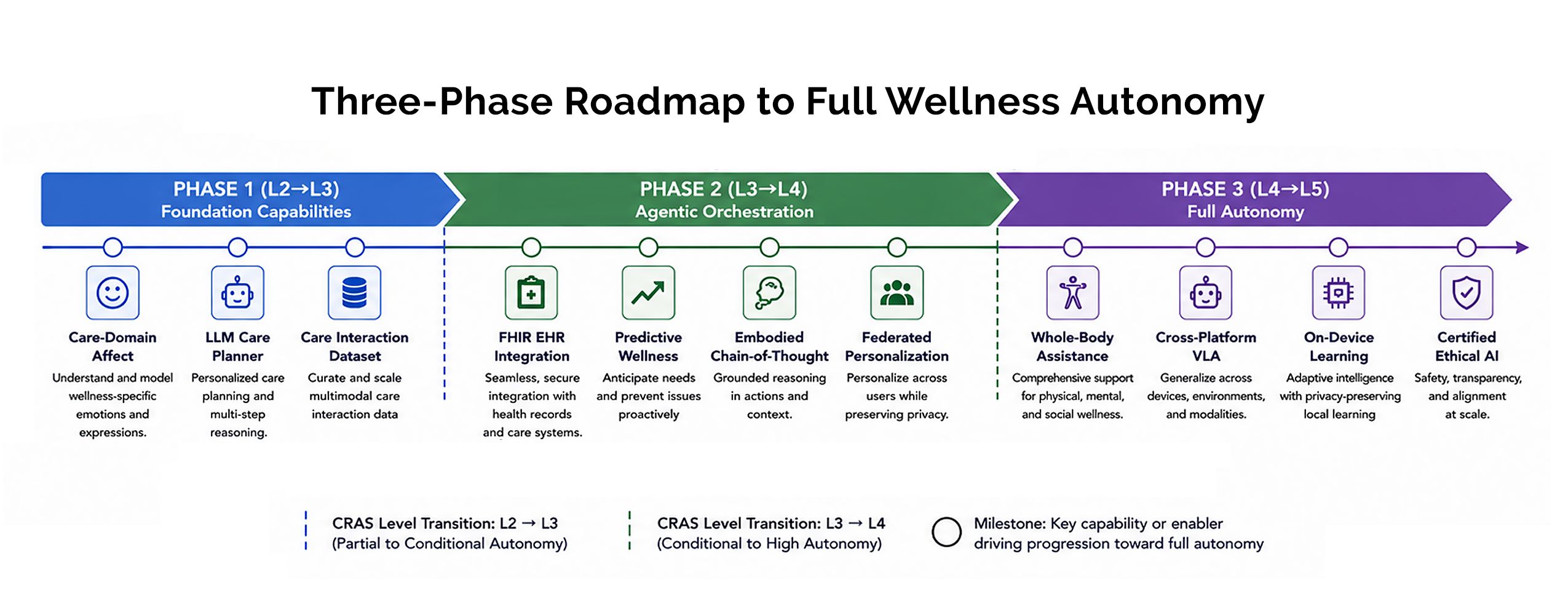}
\end{center}
\figcaption{Three-phase roadmap from current capabilities to full autonomous wellness robotics. \textbf{Phase~1} (L2$\rightarrow$L3, 2026--2028): foundation capabilities — care-domain affect, LLM care planner, care interaction dataset. \textbf{Phase~2} (L3$\rightarrow$L4, 2028--2031): agentic orchestration — FHIR EHR integration, predictive wellness, embodied chain-of-thought, federated personalization. \textbf{Phase~3} (L4$\rightarrow$L5, 2031+): full autonomy — whole-body assistance, cross-platform VLA, on-device learning, certified ethical AI. CRAS levels follow the Care Robot Autonomy Scale (L0--L5) across four dimensions: D1 Assessment, D2 Intervention, D3 Social, D4 Coordination.}
\end{figrealfull}

\begin{multicols}{2}

\subsection{Phase 1: Foundation Capabilities (CRAS L2 $\rightarrow$ L3)}

Phase~1 closes the gap between today's deployed L2 systems and a robust L3 system: condition-aware self-initiation, cross-dimensional planning, and clinical-flag escalation operating reliably in the field.

\textbf{Milestone 1: Care-domain affect recognition.} Domain-adapted facial expression and prosody models trained on older adult populations, with $\geq 80$\% balanced accuracy on a 7-affect scheme under care-environment conditions and $\leq 200$~ms inference latency. Closing the FER accuracy gap (78\% lab $\rightarrow$ 61\% care) is the single most impactful perception advance for L3.

\textbf{Milestone 2: LLM care planner with safety filter.} Goal-directed planner with structured care context injection, tool-calling against a wellness skill API, and within-session replanning. Model-independent safety post-processor evaluated against $\geq 500$ care-domain adversarial scenarios with $<1$\% harmful-output rate. Critical: \textit{model-independent} means the safety filter does not assume any particular LLM and remains effective when the underlying model is upgraded.

\textbf{Milestone 3: Care interaction dataset (CID).} A large-scale, IRB-compliant, de-identified dataset of human-robot wellness interactions with multimodal labels (affect, engagement, clinical outcome). The single most urgent collective investment in the field — without it, every research group is data-bound and every claim of clinical-grade behavior is under-evidenced. Data scaling considerations suggest $\geq 30$ care facilities and $\geq 30$ residents per ADL task to support zero-shot generalization.

\textbf{Milestone 4: Digital-twin care environments.} Simulated facility environments — Isaac Lab / MuJoCo MJX / Gazebo — with parameterized AD/ADRD resident behavioral models, allowing policies to train across thousands of scenarios. Acceptance: policies trained exclusively in twins achieve $\geq 85$\% of real-environment performance, zero-shot.

\subsection{Phase 2: Agentic Care Orchestration (CRAS L3 $\rightarrow$ L4)}

Phase~2 transforms the system from a session-aware adaptive engager into a continuous wellness-modeling care partner integrated with the clinical record.

\textbf{Milestone 5: FHIR R4 EHR integration.} Bidirectional integration with PointClickCare, MatrixCare, and Epic. Automated PHQ-9, GAD-7, SLUMS writing with proper LOINC coding; real-time clinical flag generation; care-plan reading. Acceptance: zero data integrity errors in 30-day testing; HIPAA Security Rule audit passing.

\textbf{Milestone 6: Predictive wellness modeling.} Models that forecast 30-day depression worsening, cognitive decline acceleration, and fall risk from longitudinal multimodal data. Acceptance: prospective validation AUC $\geq 0.75$ on each outcome at one or more independent sites.

\textbf{Milestone 7: Embodied chain-of-thought for clinical actions.} Plain-language pre-action rationales for every escalation, every flag, every adaptive choice that a clinician will see. Acceptance: clinician audit panel rates $\geq 90$\% of rationales as ``trustworthy and actionable'' on a defined rubric.

\textbf{Milestone 8: Federated personalization with privacy guarantees.} On-device personalization via per-resident LoRA adapters. Federated learning with secure aggregation; differential privacy on gradient updates with $\epsilon$ tuned to clinical sensitivity. Empirical Membership Inference Attack red-teaming with AUC $< 0.55$ acceptance threshold.

\subsection{Phase 3: Full Wellness Autonomy (CRAS L4 $\rightarrow$ L5)}

Phase~3 represents the frontier. Several of these milestones are not yet within the reach of any current system; we describe them so the field can engage them deliberately rather than discover them under deployment pressure.

\textbf{Milestone 9: Whole-body physical assistance with formal safety.} Sit-to-stand assistance, object retrieval, gentle physical guidance, fall-response. Requires Vision-Language-Action policies operating at 50--120~Hz, whole-body controllers in the GentleHumanoid family with tunable per-contact force limits, admittance/impedance control with care-tuned parameters, and CBF + conformal-prediction safety guarantees. Acceptance: $\geq 1{,}000$ supervised deployment hours with zero unintended forceful contacts and bounded behavioral envelope under formal verification.

\textbf{Milestone 10: Cross-platform VLA transfer.} Frontier VLA policies portable across hardware platforms via low-rank adaptation:
\[
W' = W_0 + \tfrac{\alpha}{r}\, B A, \quad B \in \mathbb{R}^{d \times r},\ A \in \mathbb{R}^{r \times k}
\]
with $r \ll d$ for parameter efficiency. Acceptance: $\geq 3$ certified hardware platforms; $<10$\% performance degradation across transfer; $\leq 4$ hours deployment configuration.

\textbf{Milestone 11: Continuous on-device learning under stability constraints.} Per-resident adaptation that is continuous yet does not drift from validated behavior. Acceptance: 6-month safety stability under continuous learning, with audit confirming no out-of-policy drift.

\textbf{Milestone 12: Certified Ethical AI.} Independent third-party audit covering: transparency of decision-making; bias evaluation across age, gender, race, ethnicity, faith, and cognitive status; resident-autonomy protection; dignity-behavior validation; clinical-boundary enforcement. Acceptance: published criteria, independently administered audit, patient-advocate panel approval before any L5 certification.

These twelve milestones, mapped to their CRAS transitions and the wellness capabilities they advance, are summarized in Table~\ref{tab:roadmap}.

\end{multicols}

\refstepcounter{table}\label{tab:roadmap}
\begin{tcolorbox}[enhanced, colback=white, colframe=dfteal, boxrule=0.6pt, arc=0pt,
  left=4pt, right=4pt, top=4pt, bottom=4pt]
{\sffamily\small\textbf{\color{dfnavy}Table 3.} Roadmap milestones, mapped to CRAS transitions and supporting wellness capability.}\\[4pt]
\sffamily\footnotesize
\renewcommand{\arraystretch}{1.3}
\begin{tabularx}{\textwidth}{@{}>{\bfseries\color{dfnavy}}p{1.6cm} >{\bfseries}p{2.6cm} X >{\bfseries\color{dfblue}}p{2.0cm}@{}}
\toprule
\textcolor{dfnavy}{\textbf{Phase / Transition}} & \textcolor{dfnavy}{Milestone} & \textcolor{dfnavy}{Acceptance criterion} & \textcolor{dfnavy}{Capability} \\
\midrule
\multirow{4}{*}{\textbf{P1: L2$\rightarrow$L3}} & Care-domain FER & $\geq 80$\% balanced acc., $\leq 200$~ms & 5.1 Sensing \\
 & LLM care planner & $<1$\% harmful on 500-scenario suite & 5.2 Planning \\
 & Care interaction dataset & $\geq 30$ facilities, $\geq 30$ residents/task & 5.1, 5.3 \\
 & Digital twin library & $\geq 85$\% sim-to-real performance & 5.2, 5.4 \\
\midrule
\multirow{4}{*}{\textbf{P2: L3$\rightarrow$L4}} & FHIR R4 EHR integration & Zero data integrity errors, HIPAA pass & 5.6 Coordination \\
 & Predictive wellness models & AUC $\geq 0.75$ on 30-day outcomes & 5.1, 5.7 \\
 & Embodied chain-of-thought & $\geq 90$\% clinician trust rating & 5.7 Explainability \\
 & Federated personalization & MIA AUC $<0.55$, $\epsilon$-DP & 5.5 Privacy \\
\midrule
\multirow{4}{*}{\textbf{P3: L4$\rightarrow$L5}} & Whole-body assist + CBF & $\geq 1{,}000$~hr, zero forceful contacts & 5.4 Safety \\
 & Cross-platform VLA & 3 platforms, $<10$\% degradation & 5.2, 5.4 \\
 & Continuous on-device learning & 6-month stability audit & 5.3, 5.4 \\
 & Certified Ethical AI & Independent audit + advocate approval & All \\
\bottomrule
\end{tabularx}
\end{tcolorbox}

\begin{multicols}{2}

\subsection{What the Field Is Missing}

Beyond the technical milestones, the field requires institutional progress that no single research group or company can provide alone:

\textbf{Standards.} CRAS-aligned certification frameworks, developed in collaboration with IEEE TC~299, ISO, ASTM, and FDA, before broad L4--L5 deployment. The standards must precede the deployment, not chase it.

\textbf{Clinical evidence base.} Multi-site randomized controlled trials of wellness robots at L3 and beyond, across diverse populations. The single-site, single-arm studies that have been the field's evidence base are insufficient for the deployment scale we anticipate.

\textbf{Reimbursement pathways.} Japan's PARO experience showed that reimbursement under the long-term care insurance scheme drives adoption far more reliably than clinical efficacy alone. CMS Innovation Center partnerships and state Medicaid HCBS waivers are the U.S.\ counterparts.

\textbf{Workforce integration.} Care staff training programs, role redefinition (especially for activities directors and nursing teams), and labor-relations engagement. Wellness robots that are perceived as replacing rather than augmenting caregivers will be rejected at the bedside, regardless of technical capability.

\subsection{Timeline}

Conservative estimates, given current research velocity and standards processes:

\begin{itemize}[leftmargin=1.2em]
\item \textbf{Phase~1 (L2$\rightarrow$L3):} 2026--2028. Several systems likely to reach robust L3 within this window. Care interaction dataset is the bottleneck; collective investment could shorten this to 2027.
\item \textbf{Phase~2 (L3$\rightarrow$L4):} 2028--2031. EHR certification and predictive validation are the long poles. FHIR pilot deployments are realistic by 2027 but full certification across major EHRs is a multi-year process.
\item \textbf{Phase~3 (L4$\rightarrow$L5):} 2031+. Whole-body physical assistance with formal safety and certified ethical AI is research at present; full deployment in Phase~3 is the target, not a forecast.
\end{itemize}

These windows are aggressive but not implausible. The technologies underlying each phase are advancing on Moore-like curves: VLA inference latency, EHR API maturity, federated learning libraries, simulation realism. The bottleneck is not the technology curve. It is the discipline of evaluation, standards, and integration with the human care system.

\end{multicols}

\begin{multicols}{2}

\section{Implications for Stakeholders}

The framework, the evidence, and the roadmap have direct implications for four stakeholder groups. We address each in turn — not as a checklist, but as a set of decisions each group must make in the next 24--36 months.

\subsection{For Care Operators}

Senior living and post-acute care operators are the immediate adopters. The first decision is not ``which robot'' but ``what CRAS level am I procuring for?'' A facility that buys an L0--L1 system expecting L3 outcomes will be disappointed. A facility that procures with eyes open at L1--L2 today, with a contracted upgrade path to L3 and L4 capabilities, can capture the available clinical and operational value while positioning for the next wave.

Concrete decisions:
\begin{itemize}[leftmargin=1.2em]
\item \textbf{Define the CRAS level} you are buying for. Match it to your infrastructure (WiFi coverage, EHR), staff readiness, and clinical oversight model.
\item \textbf{Designate a named Implementation Champion.} This person owns staff training, resident enrollment, family communication, and outcome tracking. Without one, deployments stall.
\item \textbf{Establish a baseline.} Pre-deployment PHQ-9, SLUMS, and resident-engagement metrics are essential — both for clinical evaluation and for demonstrating value to families.
\item \textbf{Treat resident data as the resident's.} The PHQ-9 score is a clinical record, not an operational metric. Operators who frame data ownership clearly avoid trust failures down the line.
\item \textbf{Plan for staff role evolution.} Activities directors become program architects; nursing teams gain a continuous data feed. The roles change; they do not disappear.
\end{itemize}

\subsubsection*{The CRAS readiness self-assessment}
The checklist in Table~\ref{tab:checklist} allows operators to evaluate readiness for each CRAS level. All criteria at the target level and below must be met before deployment. Scoring: $\geq 80$\% of target-level criteria $\rightarrow$ proceed; 60--79\% $\rightarrow$ complete outstanding items first; $<60$\% $\rightarrow$ consultation recommended.

\end{multicols}

\refstepcounter{table}\label{tab:checklist}
\begin{tcolorbox}[enhanced, colback=white, colframe=dfteal, boxrule=0.6pt, arc=0pt,
  left=4pt, right=4pt, top=4pt, bottom=4pt]
{\sffamily\small\textbf{\color{dfnavy}Table 4.} CRAS readiness self-assessment for care operators.}\\[4pt]
\sffamily\footnotesize
\renewcommand{\arraystretch}{1.3}
\begin{tabularx}{\textwidth}{@{}>{\bfseries\color{dfblue}}X *{5}{>{\centering\arraybackslash}p{1.0cm}}@{}}
\toprule
\textcolor{dfnavy}{Criterion} & \textcolor{dfnavy}{\textbf{L1}} & \textcolor{dfnavy}{\textbf{L2}} & \textcolor{dfnavy}{\textbf{L3}} & \textcolor{dfnavy}{\textbf{L4}} & \textcolor{dfnavy}{\textbf{L5}} \\
\midrule
Reliable Wi-Fi ($\geq 10$~Mbps) coverage in resident zones & \checkmark & \checkmark & \checkmark & \checkmark & \checkmark \\
Named Implementation Champion designated & \checkmark & \checkmark & \checkmark & \checkmark & \checkmark \\
Written resident consent and data-ownership protocol &\checkmark & \checkmark & \checkmark & \checkmark & \checkmark \\
Staff training completed (admin / care / IT briefings) & & \checkmark & \checkmark & \checkmark & \checkmark \\
Baseline PHQ-9 / SLUMS / engagement metrics recorded & & \checkmark & \checkmark & \checkmark & \checkmark \\
Documented clinical-oversight protocol with named clinician & & & \checkmark & \checkmark & \checkmark \\
EHR integration (FHIR R4: PCC / MatrixCare / Epic) & & & & \checkmark & \checkmark \\
Independent safety and bias audit completed & & & & \checkmark & \checkmark \\
Third-party ethics certification + patient-advocate panel & & & & & \checkmark \\
\bottomrule
\end{tabularx}
\end{tcolorbox}

\begin{multicols}{2}

\subsubsection*{Wellness robots within the care team}
Wellness robots are most effective as a complement to existing roles, changing how each role is performed rather than replacing any:
\begin{itemize}[leftmargin=1em]
\item \textbf{Activities directors} evolve from program deliverers to program architects, with a systematic analytics view of population-level wellness engagement.
\item \textbf{Nursing teams} (CRAS L3+) receive structured clinical observations directly into the EHR; daily PHQ-9 / SLUMS trends inform morning chart review.
\item \textbf{Physicians} receive a quarterly wellness summary with 90-day trend data and evidence-based care-plan recommendations.
\item \textbf{PT/OT} gain daily reinforcement of therapist-prescribed home-exercise programs, with adherence data fed back for program adjustment.
\item \textbf{Families} receive optional daily session-summary reports and contribute preference profile updates via a family portal.
\end{itemize}

\subsection{For AI and Robotics Researchers}

The research agenda implied by this framework is unusually well-defined. The framework names the gaps; the milestones (Section~8) make them measurable.

\textbf{Highest-leverage problems:}
\begin{itemize}[leftmargin=1.2em]
\item Care-domain affect recognition (Milestone~1) — foundational for L3 across the field.
\item The Care Interaction Dataset (Milestone~3) — the single most urgent collective investment, blocking nearly every downstream advance.
\item Whole-body force-adaptive control for human assistance with formal safety (Milestone~9) — the bridge from social to physical wellness robotics.
\item Long-term HRI memory architectures with bias and dignity constraints — the substrate of the relationship that drives clinical outcomes.
\item Uncertainty quantification and behavioral drift prevention in continuous-learning care systems — the safety frontier.
\end{itemize}

The senior care domain rewards research that is unusually integrative: a paper that advances perception, planning, safety, and clinical evaluation in a single system has more impact than four siloed papers each advancing one. This is itself a research-culture observation that funding agencies, conferences, and tenure committees can act on.

\subsection{For Regulators and Standards Bodies}

The senior care market is about to receive a wave of products marketed as ``AI-powered'' wellness robots. Many will deserve the label; many will not. The regulatory and standards apparatus has a brief window to set the terms of this market before market dynamics set them.

Specific recommendations:
\begin{itemize}[leftmargin=1.2em]
\item \textbf{Adopt or adapt CRAS}, or develop an equivalent autonomy framework, as the basis for evaluating wellness robot claims. SAE J3016 succeeded because it preceded large-scale autonomous vehicle deployment; the senior care field has the same opportunity now.
\item \textbf{Establish certification pathways} for L3 and L4 wellness robots, modeled on but distinct from the SaMD pathway (because wellness robots span medical and non-medical territory).
\item \textbf{Mandate transparency} of clinical-grade decisions: every flag, every escalation, every adaptive choice that has clinical implications must be auditable.
\item \textbf{Require independent ethics audit} for systems claiming L4+ capability. The audit criteria — bias, dignity, autonomy, clinical-boundary enforcement — should be published and independently administered.
\item \textbf{Support reimbursement experimentation.} The Japan PARO experience and emerging CMS Innovation Center work suggest reimbursement design is at least as important as capability development for adoption at scale.
\end{itemize}

\subsubsection*{Global context: four jurisdictions, four trajectories}
The regulatory and policy landscape varies markedly across regions; the cross-jurisdictional pattern is itself instructive.

\textbf{Japan} has the most advanced national care-robotics policy. METI has invested over \$400~million since 2013 in care-robotics development \cite{metiJapan2023}, and PARO is reimbursed under the Long-Term Care Insurance system — establishing the precedent that reimbursement, not just capability, drives adoption at scale.

\textbf{European Union.} The EU AI Act \cite{euAIAct2024} (fully applicable 2026) classifies care AI as a high-risk application, mandating conformity assessment, transparency provisions, and human oversight. Clinical features fall additionally under MDR 2017/745 (Medical Device Regulation). The combined effect is the world's most demanding regulatory regime for autonomous care systems.

\textbf{China} accounted for approximately 87\% of global humanoid robot shipments in 2025. The Fourier GR-1 platform integrated with NVIDIA's GR00T~N1 represents a credible path toward care-capable bipedal humanoids, though application to senior care specifically remains nascent.

\textbf{United States.} The FDA Digital Health Center of Excellence's evolving Software-as-a-Medical-Device guidance \cite{fdaSamd2022}, CMS Innovation Center care-coordination pilots, and state Medicaid HCBS waiver expansions together form an emerging — but still uncoordinated — framework. Cross-jurisdictional harmonization on autonomy levels, safety criteria, and clinical-evidence requirements will determine whether the global senior-care market fragments or scales.

\subsection{For Robot Platform Developers}

A note for the bipedal humanoid sector and the embodied AI ecosystem more broadly: senior care is a real opportunity, but it is not a generic deployment site for general-purpose humanoids.

The wellness robot category is defined by the wellness software stack — not by the hardware substrate. A magnificent humanoid platform without the wellness sensing, intervention library, longitudinal memory, dignity-design, and clinical integration is, in a memory care unit, an L0 system. Conversely, a well-integrated wheeled platform with the full wellness stack is L1--L2 today, with emerging L3 capabilities and a credible upgrade path beyond.

For platform developers seeking to enter senior care:
\begin{itemize}[leftmargin=1.2em]
\item \textbf{Partner rather than retrofit.} The wellness software, dataset, clinical-integration, and ethics-audit work is years of effort. Partnership with established wellness-robot teams is faster and clinically safer than building from scratch.
\item \textbf{Demonstrate force-adaptive control before claiming care capability.} GentleHumanoid-class capabilities — tunable per-contact force limits, admittance control, soft skins — are the prerequisite for deploying a generalist platform safely around frail older adults.
\item \textbf{Engage clinical evaluation early.} Outcome data from N=18 longitudinal studies is the currency that opens doors to operator pilots and regulatory acceptance. Demos do not.
\item \textbf{Treat dignity as a design constraint.} A humanoid that loops perfectly through a manipulation task while looking past the resident is failing at the actual job.
\end{itemize}

\end{multicols}

\begin{warmcallout}
\textbf{The honest assessment.} The senior care market will not be won by the most advanced humanoid, the most fluent LLM, or the most polished demo. It will be won by the system that can reliably, safely, and dignifiedly engage real residents across all seven wellness dimensions, day after day, in the conditions of an actual understaffed care facility, with outcomes that are auditable. That is a higher bar than ``AI-powered.'' It is the bar this paper has tried to specify.
\end{warmcallout}

\begin{multicols}{2}

\section{Conclusion}

This paper has argued for a category. The autonomous wellness robot is not a piece of healthcare automation, not a companion device, and not a general-purpose humanoid in a care setting. It is a system whose architecture is organized around the seven dimensions of wellness, whose autonomy is measured along the four CRAS dimensions, and whose technical capabilities — sensing, planning, personalization, safety, privacy, coordination, and explainability — are derived from the wellness paradigm itself.

We argued that the senior care crisis is structurally beyond the reach of incremental healthcare automation, and that the wellness paradigm offers a different organizing principle. We described each of the seven ICAA wellness dimensions in terms of what an autonomous robot must sense, decide, and do to support it. We defined the wellness robot category against eight specific properties, distinguishing it from neighboring categories. We introduced CRAS — a six-level autonomy framework modeled on SAE J3016, evaluated across four wellness dimensions — as the missing vocabulary for the field. We enumerated the seven technical capabilities that a system must master to climb that scale, and tied each to the wellness dimensions it supports.

We then presented Ryan as the case study: a wellness robot at CRAS Level 1--2 today, with emerging Level 3 capabilities in some dimensions, NIH/NIA-funded clinical evidence (SLUMS $+2.61$, $p{=}0.015$; PHQ-9 $-2.50$, $p{=}0.073$; 887~hours; no novelty effect), and an independent yoga study showing significant balance and cognitive improvement. Ryan is offered not as the only example but as evidence that the category is real, reproducible, and clinically meaningful.

Finally, we set out a phased roadmap of twelve technical milestones — from care-domain affect recognition through whole-body physical assistance to certified ethical AI — and identified the institutional work the field needs alongside the technical work: standards, evidence, reimbursement, workforce integration.

The senior care crisis is real, immediate, and deepening. The technology to begin addressing it is available today at CRAS Level 1--2. The technology to address it more deeply is identifiable, measurable, and within research reach. The path from current practice to full autonomy is long, but it is now mappable.

The autonomous wellness robot category is emerging. The work ahead is to give it the standards, the evidence, and the institutional infrastructure that will allow it to fulfill, at scale, the human-centered promise embedded in its name.

\end{multicols}

\begin{navycallout}
\textbf{A Call to Action.}\\[3pt]
\textit{Care operators:} Deploy at CRAS L1--2 now; every month of delay is a month residents with treatable depression and loneliness wait. Define your CRAS target; build your baseline; designate your champion; treat resident data as the resident's.\\[3pt]
\textit{Researchers:} The most urgent collective investment is the open Care Interaction Dataset. The most under-served research areas are care-domain affect recognition, whole-body force-adaptive control, and long-term HRI memory under bias and dignity constraints.\\[3pt]
\textit{Regulators \& standards bodies:} Adopt or adapt CRAS as the basis for wellness robot certification. The window to lead the market with standards rather than chase it is narrow; engage now.\\[3pt]
\textit{Robot platform developers:} The category is defined by software, data, and clinical integration — not by the hardware substrate. Partner where you can; demonstrate force-adaptive control before claiming care capability; engage clinical evaluation early.
\end{navycallout}

\vspace{0.6cm}

\vspace{4pt}
\noindent{\color{dfnavy}\sffamily\bfseries\large ACKNOWLEDGMENTS}\\[2pt]
{\color{dfteal}\rule{\linewidth}{1pt}\color{black}}\vspace{4pt}

\noindent The work described in this paper was supported by the National Institute on Aging of the National Institutes of Health under award numbers \textbf{R44AG059483} (longitudinal study of Ryan) and \textbf{SB1AG087755} (Commercialization Readiness Pilot) to DreamFace Technologies, LLC. The content is solely the responsibility of the authors and does not necessarily represent the official views of the National Institutes of Health. We are grateful to the residents, families, and staff at Eaton Senior Communities (Lakewood, CO) for their participation; to the University of Denver Ritchie School of Engineering and Computer Science; and to colleagues at the University of Texas at Arlington for their contributions to this work.

\vspace{6pt}
\noindent\textbf{Disclosure on AI-assisted editing.} The authors used Claude (Anthropic) as an editorial and formatting assistant during the preparation of this manuscript. The tool was used to support drafting, copy-editing, and document layout; it was not used to generate primary research data, clinical results, or original technical claims. All scientific content, analyses, interpretations, conclusions, and final responsibility for the work rest solely with the authors.

\vspace{8pt}
\noindent{\color{dfnavy}\sffamily\bfseries\large AUTHORS}\\[2pt]
{\color{dfteal}\rule{\linewidth}{1pt}\color{black}}\vspace{4pt}

\begin{multicols}{2}\small
\textbf{Mohammad H.\ Mahoor, Ph.D.}\\
Professor, Department of Computer Science, University of Denver; Founder, DreamFace Technologies LLC.

\textbf{Matthew Hessler}\\
Department of Computer Science, University of Denver; DreamFace Technologies LLC.

\textbf{Lombe Chileshe}\\
Department of Computer Science, University of Denver; DreamFace Technologies LLC.

\textbf{Ola Alsaadi}\\
DreamFace Technologies LLC.

\columnbreak

\textbf{Eshrat Emamian, Ph.D.}\\
Co-founder, DreamFace Technologies LLC.

\textbf{Kerstin S.\ Haring, Ph.D.}\\
Associate Professor, Department of Computer Science, University of Denver.

\textbf{Richard M.\ Voyles, Ph.D.}\\
Professor of Engineering, University of Texas at Arlington.
\end{multicols}

\vspace{8pt}
\noindent{\color{dfnavy}\sffamily\bfseries\large REFERENCES}\\[2pt]
{\color{dfteal}\rule{\linewidth}{1pt}\color{black}}\vspace{4pt}

\begin{multicols}{2}
\footnotesize
\setlength{\parskip}{2pt}
\renewcommand{\section}[2]{}

\end{multicols}

\vspace{0.4cm}
\begin{center}
{\color{dfgray}\sffamily\small
DreamFace Technologies LLC \;\textbullet\; University of Denver \;\textbullet\; University of Texas at Arlington\\
\href{https://dreamfacetech.com}{dreamfacetech.com} \;\textbullet\; May 2026}
\end{center}

\end{document}